\documentclass{ecai} 

\usepackage{latexsym}
\usepackage{amssymb}
\usepackage{amsmath}
\usepackage{amsthm}
\usepackage{booktabs}
\usepackage{enumitem}
\usepackage{graphicx}
\usepackage{color}
\usepackage{times}
\usepackage{soul}
\usepackage{url}
\usepackage[hidelinks]{hyperref}
\usepackage[utf8]{inputenc}
\usepackage[small]{caption}
\usepackage{graphicx}
\usepackage{amsmath}
\usepackage{amsthm}
\usepackage{booktabs}
\usepackage{algorithm}
\usepackage{algorithmic}
\usepackage[switch]{lineno}

\usepackage{enumitem}
\usepackage{graphicx}
\usepackage{booktabs}
\usepackage{multirow}
\usepackage{amssymb}%
\usepackage{xcolor}
\usepackage{colortbl}
\usepackage{wrapfig,lipsum,booktabs}
\usepackage{amssymb}
\usepackage{pifont}
\usepackage{subfig}
\usepackage{pifont}
\usepackage{amsmath}
\usepackage{tikz}
\newcommand{\xmark}{\ding{55}}%

\newcommand*\colourcheck[1]{%
  \expandafter\newcommand\csname #1check\endcsname{\textcolor{#1}{\ding{52}}}%
}
\colourcheck{blue}

\definecolor{airforceblue}{rgb}{0.36, 0.54, 0.66}
\definecolor{asparagus}{rgb}{0.94, 0.87, 0.8}
\definecolor{almond}{rgb}{0.53, 0.66, 0.42}
\definecolor{babyblue}{rgb}{0.54, 0.81, 0.94}
\definecolor{aureolin}{rgb}{0.99, 0.93, 0.0}

\definecolor{classname}{rgb}{0.98, 0.92, 0.84}
\definecolor{type}{rgb}{0.7, 0.75, 0.71}
\definecolor{synonym}{rgb}{0.96, 0.76, 0.76}
\definecolor{caption}{rgb}{0.54, 0.81, 0.94}
\definecolor{definition}{rgb}{0.67, 0.9, 0.93}
\definecolor{attribute}{rgb}{0.96, 0.73, 1.0}
\definecolor{track_path}{rgb}{1.0, 0.33, 0.64}

\newcommand{\BibTeX}{B\kern-.05em{\sc i\kern-.025em b}\kern-.08em\TeX}

\begin{document}


\begin{frontmatter}


\paperid{722} 


\title{TP-GMOT: Tracking Generic Multiple Object by Textual Prompt with Motion-Appearance Cost (MAC) SORT}


\author{Duy Le Dinh Anh$^{1, 2}$, Kim Hoang Tran$^{1,2}$, \textbf{Ngan Hoang Le$^2$} \\
    $^1$ FPT Software AI Center, Vietnam \\
    $^2$ Department of Computer Science, University of Arkansas, USA}


\begin{abstract}
While Multi-Object Tracking (MOT) has made substantial advancements, it is limited by heavy reliance on prior knowledge and limited to predefined categories. In contrast, Generic Multiple Object Tracking (GMOT), tracking multiple objects with similar appearance, requires less prior information about the targets but faces challenges with variants like viewpoint, lighting, occlusion, and resolution. 
Our contributions commence with the introduction of the \textbf{\text{Refer-GMOT dataset}} a collection of videos, each accompanied by fine-grained textual descriptions of their attributes. Subsequently, we introduce a novel text prompt-based open-vocabulary GMOT framework, called \textbf{\text{TP-GMOT}}, which can track never-seen object categories with zero training examples. Within \text{TP-GMOT} framework, we introduce two novel components: (i) {\textbf{\text{TP-OD}}, an object detection by a textual prompt}, for accurately detecting unseen objects with specific characteristics. (ii) Motion-Appearance Cost SORT \textbf{\text{MAC-SORT}}, a novel object association approach that adeptly integrates motion and appearance-based matching strategies to tackle the complex task of tracking multiple generic objects with high similarity. Our contributions are benchmarked on the \text{Refer-GMOT} dataset for GMOT task. Additionally, to assess the generalizability of the proposed \text{TP-GMOT} framework and the effectiveness of \text{MAC-SORT} tracker, we conduct ablation studies on the DanceTrack and MOT20 datasets for the MOT task. Our dataset, code, and models will be publicly available at: \url{https://fsoft-aic.github.io/TP-GMOT}
\end{abstract}
\end{frontmatter}

\section{Introduction}

Multiple Object Tracking (MOT) \cite{bewley2016simple} is a key task in computer vision, used for recognizing, localizing, and tracking dynamic objects in various real-world scenarios like surveillance, security, autonomous driving, robotics, and biology. Current MOT methods  \cite{bewley2016simple}, \cite{leal2016learning}, \cite{wojke2017simple}, \cite{braso2020learning}, \cite{cao2023observation}, \cite{zhang2021bytetrack}, \cite{aharon2022bot}, \cite{zeng2022motr}, \cite{du2023strongsort}, \cite{maggiolino2023deep} have limitations, relying heavily on prior knowledge, struggling with unseen categories, and having difficulty with objects of indistinguishable appearances. For example, if MOT is trained on ``car'' and ``human'' categories, it may face difficulties tracking new set of classes like ``cat'' or ``dog'' or require fine-tuning.

\begin{figure}[!t]
\centering
  \includegraphics[width=\columnwidth]{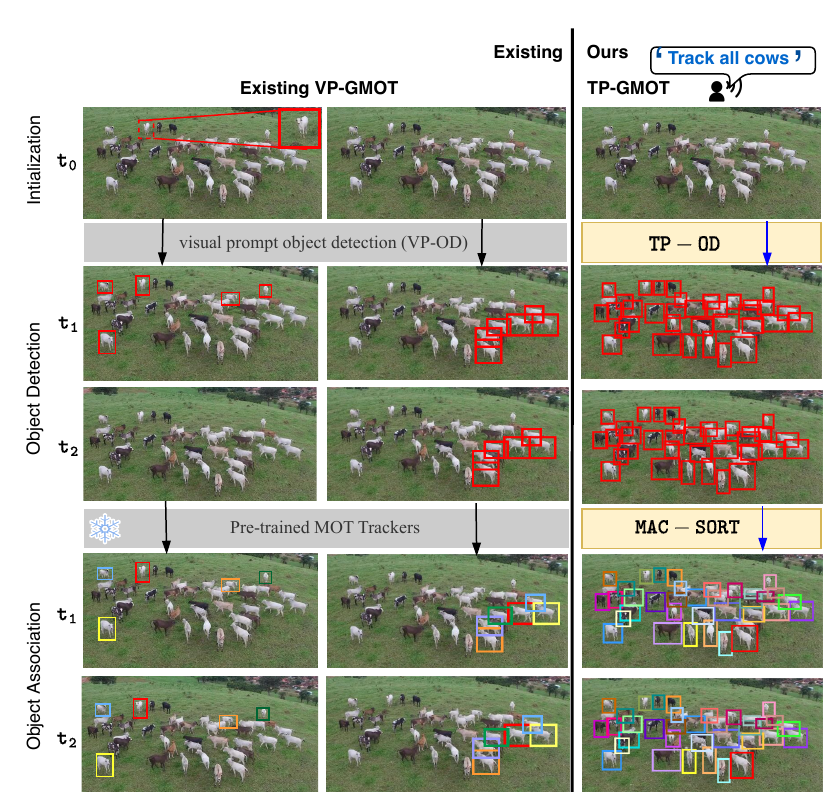}

  \caption{Comparison between our \text{TP-GMOT} and the state-of-the-art GMOT approach, {Visual Prompt-GMOT (VP-GMOT). While VP-GMOT requires an initial bounding box as input, our TP-GMOT accepts a natural language description as its input. Left: (a) VP-GMOT with two different initial bounding boxes and a MOT tracker (e.g., OC-SORT). Right: Our \text{TP-GMOT} introduces \text{TP-OD} for object detection with a natural language query and \text{MAC-SORT} for object association.} The $1^{st}$ row shows the first video frame with initialization. The $2^{nd}$ and $3^{rd}$ rows display object detection at frames $\mathtt{t_1}$ and $\mathtt{t_2}$. The $4^{th}$ and $5^{th}$ rows show object association at frames $\mathtt{t_1}$ and $\mathtt{t_2}$, respectively.}
  \vspace{1.5em}
  \label{fig:teaser}
\end{figure}

On the contrary, Generic Multiple Object Tracking (GMOT) offers a solution with less prior information about targets. GMOT aims to track multiple objects of a common or similar generic type, and is applicable in various domains, including video editing and animal behavior monitoring. However, current GMOT methods rely on visual prompt object detection (VP-OD), as detailed in \cite{luo2013generic}, \cite{luo2014bi}, and \cite{bai2021gmot}. These methods typically utilize an initial bounding box from the first video frame as a visual prompt (VP) to track all objects of the same category. This heavy reliance on the initial bounding box can lead to challenges in adapting to object variations like pose, illumination, occlusion, resolution, and texture, as shown in Fig.\ref{fig:teaser} (left).

In recent years, significant strides have been made in achieving grounded understanding through the integration of natural language processing into computer vision \cite{liu2019aligning, zhang2021consensus, li2023gligen, li2022grounded, li2023ovtrack}. This progress has enabled the precise alignment of language concepts with visual observations, allowing for a comprehensive understanding of both visual content and the nuance of natural language.

Recognizing the interest and necessity, we aim to address the limitations of both MOT and GMOT in tracking objects with specific generic attributes. To this end, we introduce a novel tracking paradigm called \textbf{\text{TP-GMOT}}, which leverages the capabilities of Vision Language Models (VLMs) to guide the tracking of multiple generic objects in videos using descriptive natural language input {as a textual prompt (TP)}. Figure \ref{fig:teaser} shows the comparison between conventional {VP-GMOT} and our proposed {\text{TP-GMOT}}. In this work, we first introduce the \textbf{\text{Refer-GMOT dataset}}, comprising various generic object categories and their corresponding fine-grained textual descriptions. {We then present \textbf{\text{TP-OD}} }(\underline{T}extual-\underline{P}rompt \underline{O}bject \underline{D}etection), a framework designed for detecting unseen objects of a category using textual prompts as input. To effectively implement TP-OD, we propose \textbf{iGDINO}, an enhancement of GroundingDINO \cite{liu2023grounding}, aimed at improving detecting objects with specific characteristics. {The proposed \text{iGDINO}} leverages a pre-trained VLM and incorporates an \textit{include-exclude (IE) strategy}, alongside a \textit{long-short memory (LSM) mechanism}. {Within the proposed \text{iGDINO}}, the IE strategy aims to mitigate false positives (FPs) issues from the pre-trained VLM. Meanwhile, the LSM mechanism handles challenges related to varying illumination, pose, and occlusion. We finally propose \text{MAC-SORT} (\underline{M}otion-\underline{A}ppearance \underline{C}ost SORT), an inventive tracking algorithm that seamlessly integrates visual appearance with motion-based matching. \text{MAC-SORT} \textit{adeptly measures appearance uniformity and dynamically balances the influence of motion and appearance during the association process}. This feature equips it to handle complex scenarios involving the tracking of objects with highly similar appearances and intricate motion patterns. Our contribution can be summarized as follows:

\begin{itemize}
\item 
We present \textbf{\text{Refer-GMOT}} dataset consisting of \text{Refer-GMOT40} and \text{Refer-Animal} datasets. They are built upon the foundations of the original GMOT-40 dataset \cite{bai2021gmot} and the AnimalTrack dataset \cite{zhang2022animaltrack} with inclusion of fine-grained natural language descriptions. 
\item We present a novel \textbf{\text{TP-GMOT}} framework, to tackle the challenges inherent in MOT and GMOT without the need of training data. {\text{TP-GMOT} follows a two-stage paradigm and introduces new contributions for each stage. In object detection, we present \text{iGDINO} to localize objects with specific characteristics of a particular type based on textual descriptions. In {object association}, we propose \textbf{MAC-SORT} to adeptly fuse visual representations with motion, enhancing the tracking of objects that have highly similar appearances and complex motion patterns.}
\item We conduct \textit{comprehensive experiments and ablation studies} on our newly introduced \text{Refer-GMOT} dataset for GMOT task. We extend our experimentation to DanceTrack \cite{sun2022dancetrack}, MOT20 \cite{dendorfer2020mot20} datasets for MOT taks, to illustrate the effectiveness and generalizability of the proposed \text{TP-GMOT} framework.
\end{itemize}

\section{Related Work}

\subsection{Multiple Object Tracking (MOT)}

Recent MOT approaches can be broadly categorized into two types based on whether object detection and association are performed by a single model or separate models, known respectively as joint detection and tracking-by-detection. In the first category \cite{CHAN2022108793}, \cite{zhou2020tracking}, \cite{pang2021quasi}, \cite{wu2021track}, \cite{yan2022towards}, \cite{meinhardt2022trackformer}, \cite{zeng2022motr}, \cite{cai2022memot}, both objects detection and objects association are simultaneously produced in a single network. In this category, object detection can be modeled within a single network with re-ID feature extraction or motion features. In the second category \cite{bewley2016simple}, \cite{leal2016learning}, \cite{wojke2017simple}, \cite{braso2020learning}, \cite{cao2023observation}, \cite{zhang2021bytetrack}, \cite{aharon2022bot}, \cite{du2023strongsort}, \cite{maggiolino2023deep} an object detector performs detecting objects in a frame, then those objects are associated with previous frame tracklets to assign identities. It is important to note that the state-of-the-art (SOTA) in MOT has been dominated by the later paradigm. \textit{Our {TP-GMOT} approach falls under the tracking by detection paradigm. Particularly, we propose {TP-OD} for zero-shot objects detector and {MAC-SORT} for associating generic objects with similar appearances.}

\subsection{Generic Multiple Object Tracking (GMOT)}

Different from MOT, which is tied to supervised learning prior knowledge and predefined categories, complicating the tracking of unfamiliar objects, GMOT \cite{luo2013generic}, \cite{luo2014bi}, \cite{bai2021gmot} aims to reduce this dependency on prior knowledge. GMOT focuses on tracking multiple objects that belong to a common or similar generic category, making it versatile for various applications, including annotation, video editing, and animal behavior monitoring. Conventional GMOT methods like \cite{luo2013generic}, \cite{luo2014bi}, and \cite{bai2021gmot} primarily follow a visual-prompt paradigm, known as VP-GMOT, where they use the initial bounding box of a single target object in the first frame to track all objects of the same class. VP-GMOT's performance can be limited due to its reliance on initial bounding boxes. It struggles when confronted with variations in object poses, lighting conditions, occlusions, and object scales. In contrast to both fully-supervised MOT and existing VP-GMOT, we introduce a novel tracking paradigm called \text{TP-GMOT}. \textit{Our \text{TP-GMOT} empowers users to track multiple generic objects in videos using natural language descriptors, eliminating the need for prior training data for predefined categories. This approach opens up new possibilities for tracking a wide range of objects with greater flexibility and adaptability, without the need for prior training data of predefined categories}.

\subsection{Pre-trained Vision-Language Models (VLMs)}

Recent computer vision tasks are trained from VLMs supervision, which has shown strong transfer ability in improving model generality and open-set recognition. CLIP \cite{radford2021learning} is one of the first works effectively learning visual representations by large amounts of raw image-text pairs. After being released, it has received a tremendous amount of attention. Some other VLMs ALIGN \cite{jia2021scaling}, ViLD \cite{gu2022open}, RegionCLIP \cite{zhong2022regionclip}, GLIP \cite{li2022grounded}, \cite{zhang2022glipv2}, UniCL\cite{yang2022unified}, X-DETR \cite{cai2022x}, OWL-ViT \cite{minderer2022simple}, LSeg\cite{li2022language}, DenseCLIP \cite{rao2022denseclip}, OpenSeg \cite{ghiasi2022open}, MaskCLIP \cite{ding2022open}, GroundingDINO \cite{liu2023grounding} have also been proposed later to illustrate the great paradigm shift for various vision tasks. Based on case studies, we can categorize VL Pre-training models into (i) image classification e.g. CLIP, ALIGN, UniCL; (ii) object detection e.g. ViLD, RegionCLIP, GLIPv2, X-DETR, OWL-ViT, GroundingDINO and (iii) image segmentation e.g. LSeg, OpenSeg, DenseSeg. The first category is based on matching between images and language descriptions by bidirectional supervised contrastive learning (UniCL) or one-to-one mappings (CLIP, ALIGN). The second category contains two sub-tasks of localization and recognition. The third group involves pixel-level classification by adapting a pre-trained language-image model. 
\textit{While Grounding DINO can detect various object categories, it faces challenges in localizing objects described by attributes. In this work, we improve Grounding DINO, an open-set object detector, to introduce iGDINO, which aims to precisely detect objects more accurately by considering their specific attributes.}

\section{Refer-GMOT dataset}
\label{sec:dataset}

\begin{table}[!thb]
\caption{Comparison of \textbf{existing datasets} of MOT, GMOT. ``\#" represents the quantity of the respective items. Cat., Vid., NLP denote Categories and Videos and Textual Language Descriptions including class\_name (cls), synonyms (syn), definition (def), attribute (att) and caption (cap).}
\setlength{\tabcolsep}{1pt}
\renewcommand{\arraystretch}{1.2}
\resizebox{\linewidth}{!}{
\begin{tabular}{c|l|c||lllll}
\toprule
& \textbf{Datasets}   & \textbf{NLP} & \#\textbf{Cat.} & \#\textbf{Vid.} & \#\textbf{Frames} & \#\textbf{Tracks} & \#\textbf{Boxs} \\ \midrule

\rowcolor{babyblue!20} 
&  MOT17~\cite{milan2016mot16}    &  \xmark   &      1        &    14      &    11.2K      &    1.3K      &   0.3M    \\ 
\rowcolor{babyblue!20}
& MOT20~\cite{dendorfer2020mot20}    &  \xmark    &        1      &     8     &     13.41K     &     3.45K     &     1.65M  \\ 
\rowcolor{babyblue!20}
& Omni-MOT~\cite{sun2020simultaneous}    &  \xmark    &          1    &     -     &      14M+    &    250K      &    110M   \\ 
\rowcolor{babyblue!20}
MOT & DanceTrack~\cite{sun2022dancetrack}  &  \xmark    &     1         &    100      &     105K     &    990      &   -    \\  
\rowcolor{babyblue!20}
& TAO~\cite{dave2020tao}   &  \xmark    &       833       &     2.9K     &    2.6M      &    17.2K      &    333K   \\
\rowcolor{babyblue!20}
& SportMOT~\cite{cui2023sportsmot}    &  \xmark    &       1       &    240      &    150K      &     3.4K     &    1.62M   \\  \cline{2-8}
\rowcolor{babyblue!50}
& \multirow{2}{*}{Refer-KITTI~\cite{wu2023referring}}   & \bluecheck (coarse)   &     \multirow{2}{*}{-}         &     \multirow{2}{*}{18}     &    \multirow{2}{*}{6.65K}      &   \multirow{2}{*}{-}       &   \multirow{2}{*}{-}    \\ 
\rowcolor{babyblue!50}
&  \multirow{-2}{*}{Refer-KITTI~\cite{wu2023referring}}   & cap only   &     \multirow{-2}{*}{2}         &     \multirow{-2}{*}{18}     &    \multirow{-2}{*}{6.65K}      &   \multirow{-2}{*}{637}       &   \multirow{-2}{*}{28.72K}   \\ \midrule \midrule
\rowcolor{aureolin!20} 
\multirow{4}{*}{GMOT}
& 
AnimalTrack~\cite{zhang2022animaltrack}  &  \xmark   &      10        &     58     &   24.7K       &   1.92K       &    429K   \\ 
\rowcolor{aureolin!20}
GMOT & GMOT-40~\cite{bai2021gmot}  &  \xmark   &        10      &    40      &    9K      &   2.02K       &  256K     \\ \cline{2-8}


\rowcolor{aureolin!40}
& \multirow{2}{*}{\texttt{Refer-GMOT}\textbf{(Ours)}} &  \bluecheck (fine-grained)   &      \multirow{2}{*}{20}        &     \multirow{2}{*}{98}     &   \multirow{2}{*}{33.7K}       &  \multirow{2}{*}{ 3.94K}      &    \multirow{2}{*}{685K}   \\
\rowcolor{aureolin!40}
& \multirow{-2}{*}{\texttt{Refer-GMOT}\textbf{(Ours)}}  &   cls, syn, def, att, cap   &  \multirow{-2}{*}{20}        &     \multirow{-2}{*}{98}     &   \multirow{-2}{*}{33.7K}       &  \multirow{-2}{*}{ 3.94K}      &    \multirow{-2}{*}{685K}    \\ 

\bottomrule
\end{tabular}}
\vspace{1.5em}
\label{tb:dataset_comparison}
\end{table}


Table \ref{tb:dataset_comparison} presents statistical information for existing tracking datasets including MOT and GMOT.
With the recent advancements and the capabilities of VLMs, there's a growing demand for including textual descriptions in tracking datasets. While natural language descriptions have already found their place in MOT datasets such as 
Refer-KITTI~\cite{wu2023referring}, they have been absent from GMOT datasets. Furthermore, textual description in Refer-KITTI ~\cite{wu2023referring} are presented as single captions, potentially limiting future explorations. Consequently, our Refer-GMOT dataset represents a pioneering effort to meet this demand by introducing fine-grained textual descriptions into the GMOT domain for the first time. It is worth noting that Refer-KITTI focuses solely on two categories, cars and humans, whereas our Refer-GMOT encompasses 20 different categories, including cars and humans. Table \ref{tb:dataset_comparison} highlights the superiority of our newly introduced Refer-GMOT over the existing Refer-KITTI from various perspectives.

The Refer-GMOT dataset comprises two subsets: Refer-GMOT40 and Refer-Animal, which utilize videos from GMOT40 \cite{bai2021gmot} and Animal \cite{zhang2022animaltrack}, respectively. Each video undergoes annotation processing, which includes the following attribute information:

\begin{figure}[!thb]
\centering
\includegraphics[width=\columnwidth]{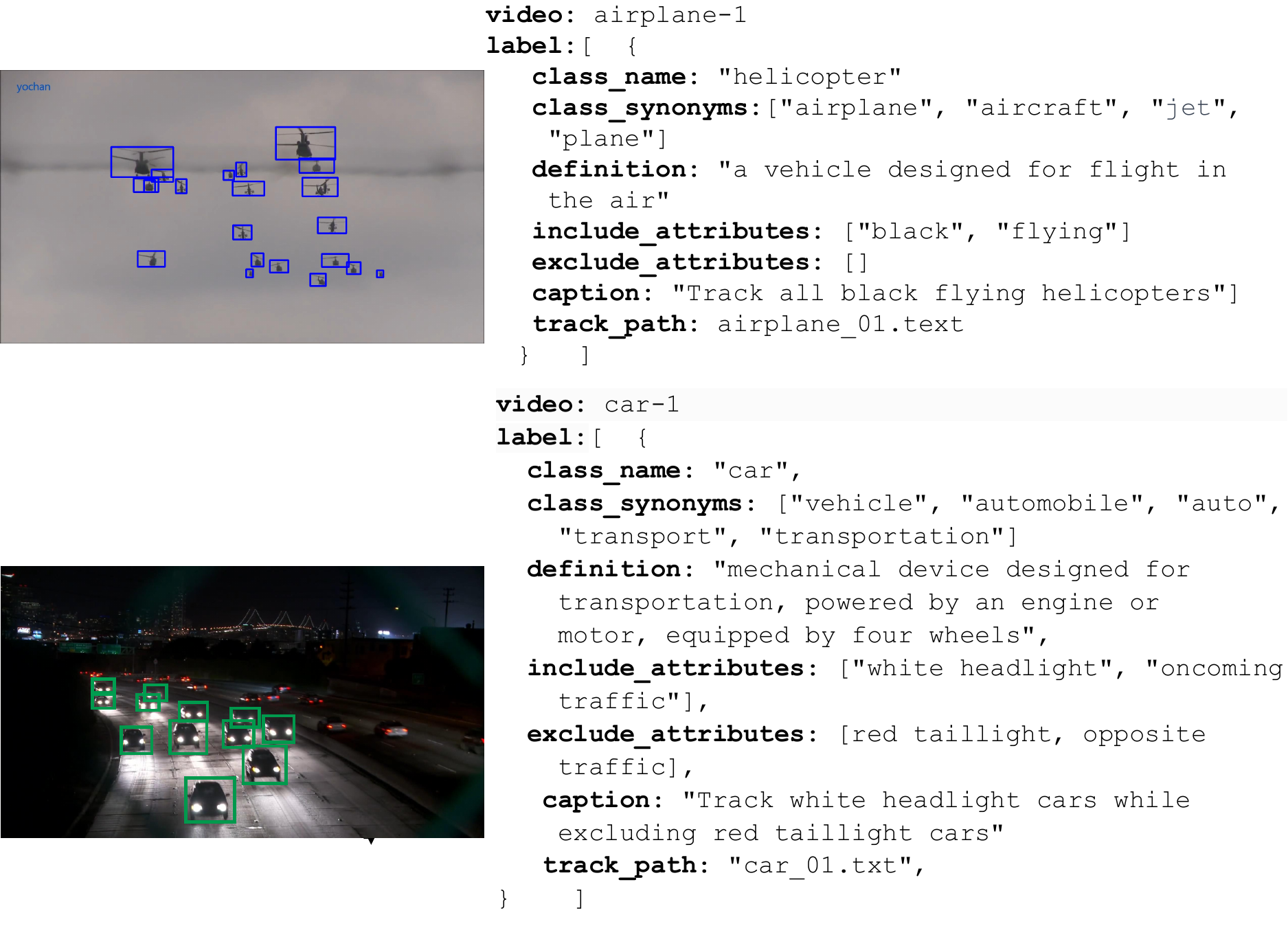}

\caption{Examples of data annotation structure in our \text{Refer-GMOT}.}
\vspace{1.5em}
\label{fig:data_example}
\end{figure}

\noindent
$\bullet$ {\texttt{class\_name}}: The generic class name of the tracked objects.

\noindent
$\bullet$ {\texttt{class\_synonyms}}: Synonyms of the class name. 

\noindent
$\bullet$ {\texttt{definition}}: A definition of the generic objects being tracked. 

\noindent
$\bullet$ {\texttt{include\_attributes}}: A list of attributes of the tracked generic objects based on visual cues. 

\noindent
$\bullet$ {\texttt{exclude\_attributes}}: A list of attributes of objects within the \text{class\_name} category that are not tracked. 

\noindent
$\bullet$ \texttt{caption}: A description of tracked generic objects. The caption for tracking all objects of the \texttt{class\_name}, follows this format: ``Track [\texttt{include\_attribute}] [\texttt{class\_name}]''. However, for situations involving the tracking of a specific subset of generic objects, the caption takes on the format: ``Track [\texttt{include\_attribute}] [\texttt{class\_name}] while excluding [\texttt{exclude\_attribute}] [\texttt{class\_name}]''. Examples of caption are given in Figure \ref{fig:data_example}.

\noindent
$\bullet$ \texttt{track\_path}: The tracking ground truth is stored in a separate file while maintaining the standardized format for MOT challenges, as outlined in \cite{milan2016mot16}, \cite{dendorfer2020mot20}. This approach ensures consistency with MOT problem conventions.

The annotation process follows the JSON format, and Figure \ref{fig:data_example} offers illustrative examples of the annotation structure. This data is conducted by 4 annotators and will be made publicly available. 



\section{Proposed \text{TP-GMOT}}
Our \text{TP-GMOT} follows the tracking-by-detection paradigm and comprises two distinct modules: textual prompt object detection (TP-OD), and object association for generic objects having high similarity appearance. In the first module, we introduce \text{iGDINO} to tackle the detection of unseen objects with specific generic attributes by textual prompt. In the second module, we present \text{MAC-SORT}, a method that leverages both appearance and motion for effective matching. 

\subsection{Proposed \text{iGDINO}}

While pre-trained VLMs have shown promise in open vocabulary object detection, they encounter certain limitations. The objects detected by these models heavily rely on threshold selection, namely, a low threshold leads to numerous FPs while a high threshold results in an increased number of FNs. While these models excel at detecting general objects such as cars, they struggle to identify objects with specific characteristics such as cars with white headlights. To address these limitations of pre-trained VLMs, {we introduce \text{iGDINO}} for detecting generic objects with specific characteristics. Leveraging GroundingDINO as a pre-trained VLM, {our \text{iGDINO}} comprises two modules as follows:

\noindent
\textbf{Module 1: Include-Exclude (IE) Strategy} 

\begin{figure}[t]
\centering
  \includegraphics[width=\columnwidth]{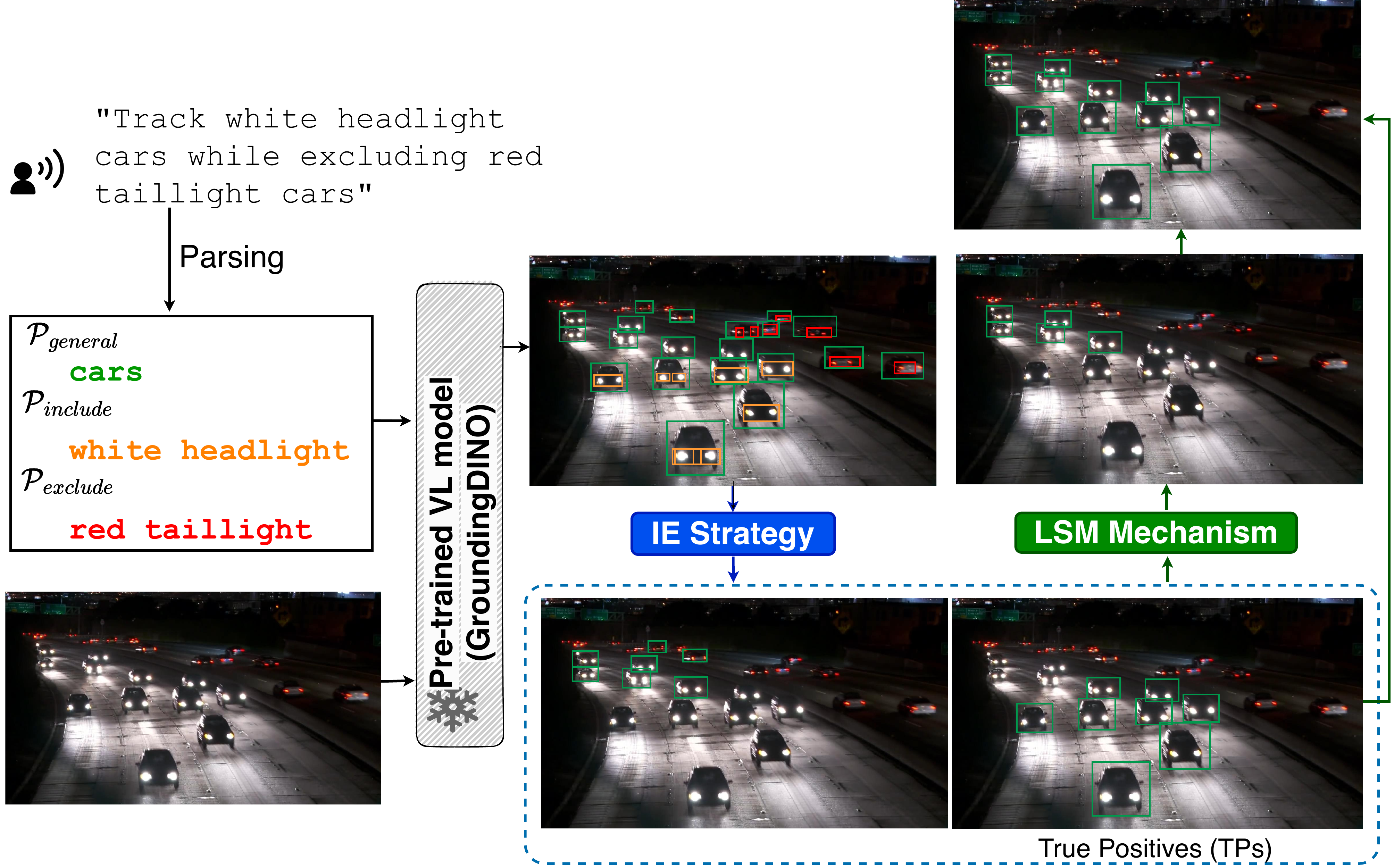}
  \caption{An illustration of our proposed \text{TP-OD} for detecting objects using input prompt. It comprises two modules of \textit{IE strategy} and \textit{LSM mechanism} to eliminate FPs from pre-trained VLM.}
\vspace{1.5em}
  \label{fig:Open-CSOD}

\end{figure}

In the typical procedure of a VLM pre-trained model, a given prompt serves as input, and in return, a set of bounding boxes is produced, each corresponding to a detected object. While VLMs have shown remarkable success in aligning textual descriptions with corresponding object bounding boxes, they still face challenges in identifying objects defined by particular attributes or when the descriptions contain negations. For example ``track all ducklings except for the mother duck''.
To precisely depict the characteristics of a generic attribute within a category, while excluding others that may belong to the same category but possess distinct traits, we introduce a novel IE strategy using the \text{caption} in the annotation defined within the proposed \text{Refer-GMOT} dataset. It's essential to note that the \text{caption} follows the format ``Track [\text{include\_attribute}] [\text{class\_name}] while excluding [\text{exclude\_attribute}] [\text{class\_name}]'' if tracking a subset of a generic type and ``Track [\text{include\_attribute}] [\text{class\_name}] if tracking all objects of a generic type. Our IE strategy is outlined as follows:

\noindent
$\bullet$\underline{Step 1:} We parse the \text{caption} into three distinct prompts: i.e., general prompt $\mathcal{P}_{general}$, include prompt $\mathcal{P}_{include}$, and exclude prompt $\mathcal{P}_{exclude}$, Herein, $\mathcal{P}_{general}$ contains [\text{class\_name}], $\mathcal{P}_{include}$ and $\mathcal{P}_{exclude}$ contain [\text{include\_attribute}] and [\text{exclude\_attribute}], respectively. In the case of tracking all objects, the $\mathcal{P}_{exclude}$ is set as $\varnothing $. For instance, in Figure \ref{fig:Open-CSOD}, the \text{caption} specifies tracking only white headlight cars and while excluding red taillight cars, thus $\mathcal{P}_{general}$ = ``\text{cars}'', $\mathcal{P}_{include}$ = ``\text{white headlight}'', 
$\mathcal{P}_{exclude}$ = ``\text{red taillight}''.

\noindent
$\bullet$ \underline{Step 2:} We process all prompts including $\mathcal{P}_{general}$, $\mathcal{P}_{include}$ and $\mathcal{P}_{exclude}$ along with a given video frame $V_i$ using a pre-trained VLM (in our case, GroundingDINO). For each prompt, we obtain a triple of ${D, F, S}$, where $D$ represents the set of bounding boxes, $F$ corresponds to their associated features and $S$ denotes their confident scores. Specifically, we obtain from general prompt: $D_g = \{d_g\}_1^{n_g}$, $F_g = \{f_g\}_1^{n_g}$, $S_g = \{s_g\}_1^{n_g}$; inclusive prompt: $D_i = \{d_i\}_1^{n_i}$, $F_i = \{f_i\}_1^{n_i}$, $S_i = \{s_i\}_1^{n_i}$; and exclusive prompt: $D_e = \{d_e\}_1^{n_e}$, $F_e = \{f_e\}_1^{n_e}$, $S_e = \{s_e\}_1^{n_e}$, where $n_g$, $n_i$ and $n_e$ are numbers of detected boxes. A particular bounding box $d_g, s_g, f_g$ is classified as:
\begin{equation}
\left\{\begin{matrix}
\text{TP} & \text{if} & \text{IoU}(d_g, d_i)\geqslant 0\\ 
\text{FP} &  \text{if} & \text{IoU}(d_g, d_e)\geqslant 0\\
\text{unclassified} &  & \text{otherwise}
\end{matrix}\right.
\end{equation}
, where IoU denotes Intersection over Union. As a result, we keep all TPs and their corresponding ${D_{tp}, F_{tp}, S_{tp}}$ and unclassified bounding boxes ${D_{u}, F_{u}, S_{u}}$ while eliminating FPs. The unclassified bounding boxes continue to be classified by the subsequent module.

\noindent
\textbf{Module 2: Long-Short Memory (LSM) Mechanism}

The unclassified boxes, denoted as ${D_{u}, F_{u}, S_{u}} = {d_u, f_u, s_u}_1^{n_u}$, with $n_u$ representing the number of unclassified boxes, undergo further processing through the proposed LSM mechanism, which involves the following steps.

\noindent
$\bullet$ \underline{Step 1:} We leverage high confidence TPs detected from previous frames and build memory bands that consist of Long Memory ($\mathcal{L}^M$) and Short Memory ($\mathcal{S}^M$). While $\mathcal{L}^M$ contains top $\kappa_1 = 9 $ highest confidence scores from the start of tracking, i.e., $\mathcal{L}^M = (D_l, F_l, S_l)$, $\mathcal{S}^M$ comprises the highest $\kappa_2 = 3$ confidence scores from the adjacent 3 frames, i.e., $\mathcal{S}^M = (D_s, F_s, S_s)$.

\noindent
$\bullet$ \underline{Step 2:} Compare the visual similarity between individual unclassified detected boxes $\{d_u, f_u, s_u\}$ with $\mathcal{L}^M$ and $\mathcal{S}^M$. 
\begin{equation}
\small
\mathtt{sim_{l-u}} = \frac{1}{\kappa_1}\sum_{i=1}^{\kappa_1}\omega(f^i_l, f_u) \text{ and } \mathtt{sim_{s-u}} = \frac{1}{\kappa_2}\sum_{i=1}^{\kappa_2}\omega(f^i_s, f_u) 
\end{equation}
, where $f^i_l \in F_l$ in $\mathcal{L}^M$ and $f^i_s \in F_s$ in $\mathcal{S}^M$. $\omega$ is cosine similarity between two vectors, i.e., $\omega(x_1, x_2) = \frac{x_1 x_2}{||x_1||||x_2||}$.

\noindent
$\bullet$ \underline{Step 3:} Compare the visual similarity between all unclassified detected boxes $\{D_u, F_u, S_u\}$, where $F_u = \{f^j_u\}_{j=1}^{n_u}$ in $\mathcal{L}^M$ and $\mathcal{S}^M$. 
\begin{equation}
\begin{split}
\mathtt{sim_{L-U}} = \frac{1}{\kappa_1 n_u}\sum_{i=1}^{\kappa_1}\sum_{j=1}^{n_u}\omega(f^i_l, f^j_u) \\
\mathtt{sim_{S-U}} = \frac{1}{\kappa_2 n_u}\sum_{i=1}^{\kappa_2}\sum_{j=1}^{n_u}\omega(f^i_s, f^j_u)
\end{split}
\end{equation}


\noindent
\underline{Step 4:} For an individual unclassified detected box $(d_{u}, f_{u}, s_{u})$, it is classified as follows:
\begin{equation}
\left\{\begin{matrix}
\text{FP} & \text{if} & \mathtt{sim_{s-u}} < \mathtt{SIM_{S-U}} \text{  and   }\mathtt{sim_{l-u}} < \mathtt{SIM_{L-U}}\\ 
\text{TP} &  & \text{otherwise}
\end{matrix}\right.
\end{equation}

\begin{figure}[t]
\centering
  \includegraphics[width=\columnwidth]{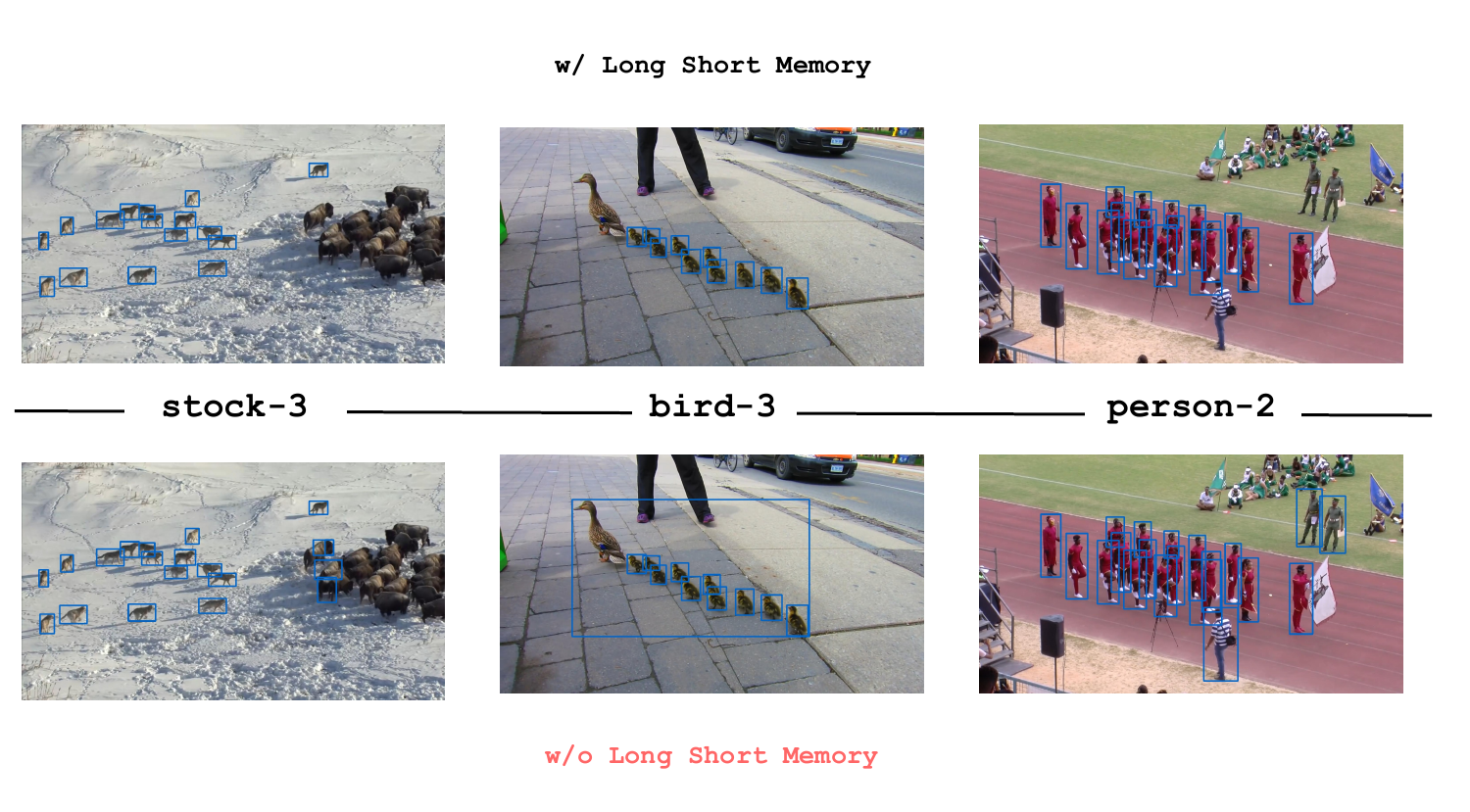}
  \caption{Comparison between apply \textit{LSM mechanism} and not to eliminate FPs from pre-trained VLM.}
\vspace{1.5em}
  \label{fig:LSM}

\end{figure}

For all unclassified detected boxes, which are determined as TPs, are combined with TPs from the Module 1 - IE strategy to create the final TPs set as demonstrated in Figure \ref{fig:Open-CSOD}. 


\subsection{Proposed \text{MAC-SORT}}


Let us begin by outlining the problem settings as follows: Let $\mathcal{T} \in \mathbb{R}^{N\times 7}$ denote a set of $N$ existing tracks, with each track defined as a vector of $[u,v,s,r,\bar{u},\bar{v},\bar{s}]^T$. Let $\mathcal{D}\in \mathbb{R}^{M\times 5}$ denote a set of $M$ detections in the new-coming time step, where each detected bounding box $M_i$ is represented by two vectors, namely, a position vector of $[u, v, w, h, c]^T$ and an embedding vector $f_i$. The variables $(u, v)$ correspond to the 2D coordinates of the object center. $s$ represents the bounding box scale (area), and $r$ denotes the bounding box aspect ratio, assumed to be constant. The other variables, $(\bar{u},\bar{v},\bar{s})$, represent the corresponding time derivatives. $w$ and $h$ denote the width and height of the bounding box, with detection confidence $c$.

The standard similarity between track and box embeddings is defined using cosine distance, denoted as $C_a \in \mathbb{R}^{M\times N}$. In a typical tracking approach that combines visual appearance and motion cues, the cost matrix $C$ is computed as $C(\mathcal{T}, \mathcal{D}) = M_c(\mathcal{T}, \mathcal{D}) + \alpha C_a(\mathcal{T}, \mathcal{D})$, where $M_c$ represents the motion cost, measured by the IoU cost matrix. Leveraging OC-SORT, which computes a virtual trajectory over the occlusion period to rectify the error accumulation of filter parameters during occlusions, the motion cost is defined as $M_c(\mathcal{T}, \mathcal{D}) = IoU(\mathcal{T}, \mathcal{D}) + \lambda C_v(\tilde{T}, \mathcal{D})$. Thus, the resulting cost matrix integrating both visual appearance and motion is as follows:
\begin{equation}
C(\mathcal{T}, \mathcal{D}) = IoU(\mathcal{T}, \mathcal{D}) + \lambda C_v(\tilde{T}, \mathcal{D}) + \alpha C_a(\mathcal{T}, \mathcal{D})
\label{eq:cost_0}
\end{equation}
{where $\tilde{T}$ contains the trajectory of observations of all existing tracks.} The velocity cost \( C_v \) in Equation 6 measures the angular consistency between the observed motion direction of a track and its intended direction towards a new detection. Specifically, the angular difference \( \Delta \theta \) is calculated as:

\[
C_v = \Delta \theta = \left| \arctan\left(\frac{v_1 - v_2}{u_1 - u_2}\right) - \arctan\left(\frac{v_3 - v_4}{u_3 - u_4}\right) \right|
\]

where \((u_1, v_1) \in \mathcal{T}^{t-2}\) and \((u_2, v_2) \in \mathcal{T}^{t-1}\) are the coordinates of the previous observations forming the track direction, and \((u_3, v_3) \in \mathcal{T}\) and \((u_4, v_4) \in \mathcal{D}\) are the coordinates forming the direction towards the new detection.

To strike a balance between visual appearance and motion cues, we incorporate adaptive appearance weight $W_{aaw}$ and adaptive motion cost $W_{amc}$ into Equation \ref{eq:cost_0}, resulting in the rewritten form:
\begin{equation}
C(\mathcal{T}, \mathcal{D}) = W_{amc}IoU(\mathcal{T}, \mathcal{D}) + \lambda C_v(\tilde{T}, \mathcal{D}) + W_{aaw}C_a(\mathcal{T}, \mathcal{D})
\label{eq:cost_1}
\end{equation}
To effectively handle the high similarity between objects of the same generic type in GMOT, we propose the following hypothesis: when the visual appearances of all detections are very similar, the tracker should prioritize motion over appearance. The homogeneity of visual appearances across all detections can be quantified as follows:
\begin{equation}
\mu = \frac{1}{M}\sum_{i=1}^{M}{f_i} \text{ and }
\mu_{det} = \frac{1}{M}\sum_{i=1}^{M}{\mathtt{cos}(f_i, \mu)}
\label{eq:distribution}
\end{equation}
\noindent
, where $M$ represents the number of detections, and $f_i$ represents the feature of detection $i$. Here, we consider a threshold $\theta$ to determine the similarity between two vectors; if the angle between them is smaller than $\theta$, the vectors are considered more similar. It's noteworthy that when $\mu_{det} > \cos(\theta)$, the visual appearance is less reliable for tracking, implying that $W_{aaw}$ should be less than 1. Conversely, $W_{aaw} > 1$ when $\mu_{det} < \cos(\theta)$. Thus, the weight $W_{aaw}$ is defined as:
\begin{equation}
    W_{aaw} = \frac{1 - \mu_{det}}{1 - \mathtt{cos}(\theta)}
\label{eq:theta}
\end{equation}
{We initialize the motion weight at 1 and incorporate its complement, calculated as (1 - $W_{aaw}$), to intensify the emphasis on motion cost. This adjustment ensures that as the similarity between objects increases, the tracker increasingly prioritizes motion in its tracking decisions.} Thus, the adaptive motion weight $W_{amc}$ is defined as:
\begin{equation}
    W_{amc} = 1 + (1 - W_{aaw}) = 2 - \frac{1 -\mu_{det}}{1-\mathtt{cos}(\theta)}
\end{equation}

As a result, the final cost matrix $C$ is computed as follows:
\begin{equation}
\begin{aligned}
C(\mathcal{T}, \mathcal{D}) = & \left(2 - \frac{1 - \mu_{det}}{1-\mathtt{cos}(\theta)}\right)IoU(\mathcal{T}, \mathcal{D}) \\ 
+ & \lambda C_v(\tilde{T}, \mathcal{T}) 
+  \frac{1 - \mu_{det}}{1 - \mathtt{cos}(\theta)}C_a(\mathcal{T}, \mathcal{D})
\end{aligned}
\label{eq:cost_1}
\end{equation}

\section{Experiments}
\noindent
\textbf{Datasets.}
For performance evaluation, we mainly benchmark our \text{TP-GMOT} on the proposed \text{Refer-GMOT} dataset consisting of two subsets of Refer-GMOT40 and Refer-Animal for the GMOT task. To demonstrate the generalizability of \text{TP-GMOT} framework and the effectiveness of \text{MAC-SORT} tracker, we extend our evaluation to include \textit{DanceTrack} \cite{sun2022dancetrack} and \textit{MOT20} \cite{dendorfer2020mot20} datasets for the MOT task. The \text{Refer-GMOT} dataset, consisting of \text{Refer-GMOT40} and \text{Refer-Animal} dataset, is described in Section \ref{sec:dataset}. \textit{DanceTrack} is a vast dataset designed for multi-human tracking i.e., group dancing. It includes 40 train, 24 validation, and 35 test videos, totaling 105,855 frames recorded at 20 FPS. \textit{MOT20} is an updated version of MOT17 \cite{milan2016mot16} including more crowded scenes, object occlusion, and smaller object size than MOT17.

We employ the following metrics: Higher Order Tracking Accuracy ($HOTA$) \cite{Luiten_2020hota}, Multiple Object Tracking Accuracy ($MOTA$) \cite{Bernardin2008mota}, $IDF1$ \cite{Ristani2016idf1}. $HOTA$ and $MOTA$ are standard metrics for MOT, while $IDF1$ serves as an ID metric. It's important to note that $HOTA$ is measured based on Detection Accuracy ($DetA$), and overall Association Accuracy $AssA$, i.e. $HOTA = \sqrt{DetA \cdot AssA}$, thus, it effectively strikes a balance in assessing both frame-level detection and temporal association performance.

We compare our method with previous SOTA methods, including one-stage paradigm:  Siamese-DETR~\cite{liu2023siamesedetr}, OVTrack~\cite{li2023ovtrack} and two-stage paradigm:  SORT~\cite{bewley2016simple}, DeepSORT~\cite{wojke2017simple}, ByeTrack~\cite{zhang2021bytetrack}, MOTRv2~\cite{zhang2023motrv2}, VP-SORT~\cite{cao2023observation}, Deep-OCSORT~\cite{maggiolino2023deep}, TransTrack\cite{transtrack}, QDTrack\cite{fischer2023qdtrack}, MOTDT~\cite{Chen2018RealTimeMP}, MeMOT~\cite{cai2022memot}, FairMOT~\cite{zhang2021fairmot}, GSDT~\cite{Wang2020_GSDT}, CSTrack~\cite{liang2022cstrack}.

\textbf{Implementation detail.}

In the implementation of iGDINO, we adopt the settings from  GroundingDINO \cite{liu2023grounding}, employing Swin-B as the backbone, and apply a threshold of 0.2 to filter the detection results. Regarding the $\lambda$ hyperparameter in MAC-SORT, we set it to a constant value of 0.2, consistent with the default configuration in OC-SORT. All experiments and comparisons have been conducted by an NVIDIA A100-SXM4-80GB GPU.

\begin{table}[!t]
\setlength{\tabcolsep}{2pt}
\renewcommand{\arraystretch}{1.2}
  \centering
  \caption{\textit{GMOT Tracking comparison} on \textit{Refer-GMOT40} dataset between our {\textbf{TP-OD}} with existing SOTA detectors GlobalTrack \cite{bai2021gmot} and GLIP on various SOTA trackers. VP and TP denote visual prompt and textual prompt, respectively.}

\label{tab:gmot40}
\resizebox{\columnwidth}{!}{
\begin{tabular}{l|l|lc|ccc}
\toprule
\textbf{Paradigm} &\textbf{Trackers}   & \textbf{Detectors} & \textbf{Prompt} &  \textbf{HOTA}$\uparrow$ & \textbf{MOTA}$\uparrow$ & \textbf{IDF1}$\uparrow$ \\ \midrule
\multirow{2}{*}{One Stage} & Siamese-DETR & Siamese-DETR & VP & - & 50.00 & 51.30 \\ 
& OVTrack & OVTrack & TP & 34.13 & 27.78 & 36.07 \\ \hline
\multirow{24}{*}{Two Stage} 

&\multirow{4}{*}{SORT}   &   GlobalTrack & VP                   & 30.05          & 20.83             &   33.90           \\ 
 & &GLIP & TP & 53.08          & 51.61   & 60.01 \\
  & & GroundingDINO & TP &  \textbf{56.97}  & 64.82  & 67.50 \\
 & & \textbf{{iGDINO} (Ours)}& TP  & 56.43 &   \textbf{66.72} &  \textbf{67.51} \\ \cline{2-7}
 
&\multirow{3}{*}{DeepSORT}        & GlobalTrack & VP   & 27.82          & 17.96     & 30.37   \\  
 & &GLIP & TP & 49.31          & 44.84       & 53.95 \\
   & & GroundingDINO & TP &  \textbf{52.08}  & 57.22  & \textbf{59.98} \\
  & & \textbf{{iGDINO} (Ours)}& TP  & 50.54 &  \textbf{60.21}  & 57.93 \\ \cline{2-7}

&\multirow{3}{*}{ByteTrack}       & GlobalTrack  & VP   & 29.88          & 20.29     & 34.70  \\
& &GLIP & TP & 52.84     & 50.93    & 62.36 \\
  & & GroundingDINO & TP &  \textbf{56.48}  & 62.89  &  \textbf{69.80} \\
 &   & \textbf{{iGDINO} (Ours)} & TP  & 55.87 & \textbf{64.79} & 69.79 \\ \cline{2-7}
 
& \multirow{3}{*}{MOTRv2}     & GlobalTrack  & VP   & 23.76 & 13.87 & 25.17 \\ 
& & GLIP & TP &31.32 & 18.54 & 31.28 \\
  & & GroundingDINO & TP & \textbf{33.18}   & 18.65  & 32.37 \\
 & & \textbf{{iGDINO} (Ours)}& TP  & 32.93 & \textbf{18.70} &  \textbf{33.48 }\\\cline{2-7}
 
& \multirow{3}{*}{OC-SORT}        & GlobalTrack  & VP   & 29.00 &  19.96 &  32.85 \\ 
 & &GLIP & TP & 55.20   & 52.51      & 63.33 \\
   & & GroundingDINO & TP &  \textbf{57.68}  &  62.94 & 67.50  \\
  &  & \textbf{{iGDINO} (Ours)}& TP  & 56.06 &  \textbf{63.69} & \textbf{68.85} \\ \cline{2-7}
  
& \multirow{3}{*}{Deep-OCSORT}       & GlobalTrack  & VP   & 30.37 & 21.10 & 34.74  \\
& &GLIP & TP & 54.19     & 52.28    & 61.41 \\
  & & GroundingDINO & TP &  \textbf{57.81}  &  65.08 &  \textbf{68.55} \\
 & & \textbf{{iGDINO} (Ours)}& TP  & 55.74 & \textbf{65.53} & 66.54  \\ 
 
\bottomrule  
\end{tabular}
}

\label{tb:comp_csod}
\end{table}

\begin{table}[!t]
  \centering
  \caption{\underline{GMOT Tracking comparison} on \textit{Refer-GMOT40} dataset between our {\textbf{MAC-SORT}} with various SOTA trackers on the same our iGDINO detector.}

  \label{tab:gmot40}
  \resizebox{0.85\columnwidth}{!}{%
\begin{tabular}{l|ccc}
\toprule
\textbf{Trackers}   &  \textbf{HOTA}$\uparrow$ & \textbf{MOTA}$\uparrow$ & \textbf{IDF1}$\uparrow$ \\ \midrule
SORT   & \underline{56.43} &   \underline{66.72} &  67.51  \\ 
DeepSORT     & 50.54 &  {60.21}  &  57.93  \\ 
ByteTrack  & 55.87 & 64.79 & \underline{69.79}  \\ 
OC-SORT     & 56.06 &  63.69 &  68.85  \\ 
Deep-OCSORT  & 55.74 & {65.53} &  66.54  \\ 
MOTRv2  &    32.93 & 18.70 &  33.48 \\ \midrule
{\textbf{MAC-SORT}}  (\textbf{Ours})&   \textbf{58.58} & \textbf{67.77} & \textbf{71.70} \\ 
\bottomrule
\end{tabular}}
 \vspace{1.5em}
\label{tb:comp_macsort-gmot}
\end{table}

\begin{table}[!t]
  \centering
  \caption{\underline{GMOT Tracking comparison} on \textit{Refer-Animal} between our \textbf{{MAC-SORT}} and \textbf{{TP-OD}} with existing \textit{fully-supervised MOT} methods. These methods utilize Faster R-CNN (FRCNN) and YOLOX trained on the AnimalTrack-trainset as their object detector. It is important to note that these fully-supervised methods are limited in their ability to handle category-agnostic tracking.}
  \resizebox{1.0\linewidth}{!}{
    \begin{tabular}{l|l|c|ccc}
    \toprule
    \multirow{2}{*}{\textbf{Trackers}}                    & \multirow{2}{*}{\textbf{Detectors}}    & \textbf{Category}                  & \multirow{2}{*}{\textbf{HOTA}$\uparrow$}                  & \multirow{2}{*}{\textbf{MOTA}$\uparrow$}                  & \multirow{2}{*}{\textbf{IDF1}$\uparrow$}    \\
    & &\textbf{Agnostic} & & & 
    \\ \midrule

    \multirow{1}{*}{SORT}      & FRCNN  & \xmark & 42.80                           & 55.60                           & 49.20                   \\
    \multirow{1}{*}{DeepSORT}  & FRCNN  & \xmark & 32.80                           & 41.40                           & 35.20                                         \\
    \multirow{1}{*}{ByteTrack} & YOLOX       & \xmark & 40.10                           & 38.50                           & 51.20                                                   \\ 
    \multirow{1}{*}{TransTrack} & YOLOX        & \xmark & 45.40                           & 48.30                           & 53.40                                                   \\
    \multirow{1}{*}{QDTrack} & YOLOX        & \xmark & 47.00                           & 55.70                           & 56.30      \\  
    \multirow{1}{*}{OC-SORT} & YOLOX        & \xmark &  56.93 & 65.02 & \underline{67.48} \\   
    \multirow{1}{*}{Deep-OCSORT} & YOLOX        & \xmark &  57.24 & \underline{68.05} & 62.01 \\  
    \multirow{1}{*}{{MOTRv2}} & YOLOX        & \xmark & 52.07 & 57.08 & 62.07 \\  \midrule 
    \textbf{{MAC-SORT}(Ours)}&  YOLOX     & \xmark &   \textbf{57.86}   &  \textbf{68.32}   &   63.01       \\    
    \textbf{{MAC-SORT}(Ours)}&  \textbf{{TP-OD}} \textbf{(Ours)}      & \bluecheck &      \underline{57.29}       &   66.46   &  \textbf{68.37}    \\ \bottomrule              
    \end{tabular}
}
  \label{tb:Animal-ZGMOT}
\label{tb:comp_macsort-animal}
\end{table}%

\subsection{Performance Comparison.}
Table \ref{tb:comp_csod} presents a benchmark comparison of GMOT task across two paradigms on Refer-GMOT40 dataset. In the one-stage tracking paradigm, we include Siamese-DETR  utilizing visual prompt (VP) and OVTrack employing textual prompt (TP). In the two-stage tracking paradigm, we utilize GlobalTrack \cite{huang2020globaltrack} for VP-OD, while for TP-OD, we include GLIP, GroundingDINO, and our introduced iGDINO.
In our experimental setup, VP-OD undergoes five independent runs, whereas TP-OD leverages captions derived from the dataset. Comparing VP-OD to TP-OD, it is clear that TP-OD outperforms with notable gaps across metrics. Within TP-OD, our proposed iGDINO demonstrates advantages compared to GLIP and GroundingDINO. Thus, we will leverage iGDINO as the object detector to benchmark trackers.


 
Table \ref{tb:comp_macsort-gmot} shows a comparison between \text{MAC-SORT} with various trackers using the same object detector iGDINO on \text{Refer-GMOT40} dataset. It is evident that \text{MAC-SORT} consistently outperforms other trackers. In comparison to similar methods DeepSORT and Deep-OCSORT that incorporate both visual cues and motion, \text{MAC-SORT} also demonstrates superior performance across all metrics.

Table \ref{tb:comp_macsort-animal} presents a comparison between our proposed \text{MAC-SORT} and other existing fully-supervised MOT methods on the \text{Refer-Animal} dataset. Those methods utilize Faster R-CNN (FRCNN) \cite{ren2015faster} and YOLOX \cite{yolox2021} trained on AnimalTrack-trainset as their object detector. In order to ensure a fair comparison, we have also implemented a fully-supervised \text{MAC-SORT} method with YOLOX object detection. While our \text{MAC-SORT} with YOLOX object detector achieves the best performance on both HOTA, MOTA, it is worth noting that TP-OD, which does not require any training data, surpasses majority fully-supervised MOT methods.

\begin{table}[!ht]
\centering
\setlength{\tabcolsep}{4pt}
\renewcommand{\arraystretch}{1.0}
\caption{\underline{Ablation study of generalizability} of \textbf{\texttt{TP-GMOT}} framwork on \textit{DanceTrack-valset} with \underline{MOT task}. We compare our \textbf{\texttt{MAC-SORT}} and \textbf{\texttt{TP-OD}} with other \textit{fully-supervised MOT} methods, which use YOLOX trained on DanceTrack-trainset as their object detector. \textcolor{gray}{Deep-OCSORT} denotes the reported results in the paper whereas Deep-OCSORT$^\dag$ presents the reproduced results with the best settings suggested by the authors on our machine with the same object detector. It is important to note that these existing fully-supervised methods are limited in their ability to handle category-agnostic tracking.}
    \resizebox{1.0\linewidth}{!}{\begin{tabular}{l|l|c|ccc}
    \toprule
    \multirow{2}{*}{\textbf{Trackers}}                    & \multirow{2}{*}{\textbf{Detectors}}    & \textbf{Category}                  & \multirow{2}{*}{\textbf{HOTA}$\uparrow$}                  & \multirow{2}{*}{\textbf{MOTA}$\uparrow$}                  & \multirow{2}{*}{\textbf{IDF1}$\uparrow$}  \\
    & &\textbf{Agnostic} & & &
    \\ \midrule
    \multirow{1}{*}{SORT}    & YOLOX                      & \xmark                      & 47.80   & \textbf{88.20}    & 48.30   \\
    \multirow{1}{*}{{DeepSORT}}   & YOLOX                    & \xmark                      & 45.80    & 87.10    & 46.80  \\
    \multirow{1}{*}{MOTDT}  & YOLOX                 & \xmark                      &   39.20   &   84.30   &    39.60 \\
    \multirow{1}{*}{ByteTrack} & YOLOX                 & \xmark                      & 47.10     & \textbf{88.20}    & 51.90  \\ 
    \multirow{1}{*}{OC-SORT}  & YOLOX                 & \xmark                      &  52.10    &   \underline{87.30}   &    51.60   \\ 
 \textcolor{gray}{Deep-OCSORT} & \textcolor{gray}{YOLOX} & \textcolor{gray}\xmark & \textcolor{gray}{58.53} & \textcolor{gray}{--} & \textcolor{gray}{59.06}  \\
     Deep-OCSORT$^\dag$ & YOLOX & \xmark & 49.36 & 84.82 & 48.89 \\
    \bottomrule
    \multirow{1}{*}{\textbf{\texttt{MAC-SORT}}}\textbf{(Ours)} & YOLOX                 & \xmark                  &   \textbf{53.78} & 86.85 & \textbf{54.06}   \\                               
   \textbf{\texttt{MAC-SORT}(Ours)}&  \textbf{\texttt{TP-OD}} \textbf{(Ours)} & \bluecheck                      & 48.75 & 81.74 & 48.02   \\ \bottomrule

    \end{tabular}
}
\vspace{1.5em}
    \label{tab:Abl_DanceTrack}
\end{table}

\begin{table}[!t]
\setlength{\tabcolsep}{4pt}
\renewcommand{\arraystretch}{0.9}
  \centering
  \caption{\underline{Ablation study on the effectiveness of \texttt{MAC-SORT}} on \textit{MOT20-testset} with \underline{MOT task}. We compare our \textbf{\texttt{MAC-SORT} } with other SORT-based \textit{MOT} methods. As \textcolor{gray}{ByteTrack, OC-SORT} uses different thresholds for testset sequences and offline interpolation procedure, we also report scores by disabling these as in ByteTrack$^\dag$, OC-SORT$^\dag$. As \textcolor{gray}{Deep OC-SORT} used separated weights for YOLOX object detector, we also report scores by retraining YOLOX on MOT20-trainset as in Deep OC-SORT$^\dag$.}
  \label{tab:gmot40}
  \resizebox{0.75\columnwidth}{!}{%
\begin{tabular}{l|ccc}
\toprule
\textbf{Trackers}  &  \textbf{HOTA}$\uparrow$ & \textbf{MOTA}$\uparrow$ & \textbf{IDF1}$\uparrow$  \\ \midrule
MeMOT  & 54.1 & 63.7 & 66.1        \\  
 FairMOT & 54.6 & 61.8 & 67.3         \\   
 GSDT    & 53.6 & 67.1 & 67.5    \\  
  CSTrack     & 54.0 & 66.6 &  68.6  \\  
 \color{gray}{ByteTrack}      & \color{gray}61.3 & \color{gray}{77.8}  & \color{gray}75.2  \\ 
 \color{gray}OC-SORT    &  \color{gray}62.4 &	\color{gray}75.7 &	\color{gray}76.3  \\ 
  \color{gray}{Deep-OCSORT}  & \color{gray}63.9 &\color{gray}75.6   &\color{gray}79.2  \\ 
 ByteTrack$^\dag$     & 60.4 & 74.2  & 74.5 \\ 
 OC-SORT$^\dag$   & \underline{60.5} & 73.1  & 74.4 \\ 
  Deep OC-SORT$^\dag$ &59.6 & \textbf{75.3} &\underline{75.2} \\ \midrule
 {\textbf{\texttt{MAC-SORT}} \textbf{(Ours)}} & \textbf{62.6} & \underline{75.2}  & \textbf{76.9} \\
\bottomrule
\end{tabular}}
\vspace{1.5em}
\label{tb:Abl_MOT20}
\end{table}

\subsection{Computation Efficiency}
\begin{table}[!h]
\centering
\caption{Computational cost and performance comparison on \textit{Refer-GMOT40} dataset between GLIP, GroundingDINO and our \textbf{iGDINO}.}
\resizebox{\linewidth}{!}{
\begin{tabular}{ll|ccc|cc}
\toprule
\multicolumn{2}{c|}{\textbf{Tracking Methods}} & \multicolumn{3}{c|}{\textbf{Performance}} & \multicolumn{2}{c}{\textbf{Computation}} \\
\midrule
\textbf{Detectors} &  \textbf{Trackers} & \textbf{AP$_{50}\uparrow$} & \textbf{HOTA}$\uparrow$ & \textbf{MOTA}$\uparrow$ & \textbf{FPS}$\uparrow$ & \textbf{Params}$\downarrow$ \\
\midrule
\multirow{2}{*}{GLIP} & SORT & \multirow{2}{*}{66.2} & 53.08 & 51.61 & 3.52  &  \multirow{2}{*}{232M}  \\
& \textbf{\texttt{MAC-SORT}} & & 54.91 & 52.83 & 3.40 &  \\ \midrule
GroundingDINO & \textbf{\texttt{MAC-SORT}} & 71.7 & 58.31 & 65.60 & \textbf{6.10} & \textbf{172M} \\ \midrule
\textbf{\texttt{iGDINO}} & \textbf{\texttt{MAC-SORT}} & \textbf{72.7} & \textbf{58.58} & \textbf{67.77} & 6.02 & \textbf{172M} \\
\bottomrule
\end{tabular}
}
\label{tab:compute_cost}
\end{table}

We analyze the network computation complexity versus tracking performance in Table \ref{tab:compute_cost}, which includes GLIP, GroundingDINO, and iGDINO as TP-OD, along with SORT and MAC-SORT as trackers. By comparing different trackers on the same TP-OD or comparing various TP-OD with the same tracker, we observe that the combination of iGDINO and MAC-SORT delivers the best performance with lower computational complexity.


\subsection{Ablation Study}

\noindent
\underline{Generalizability of \text{TP-GMOT} framework on MOT task.} In addition to the GMOT task, we also evaluate its generalizability on the MOT task, as in Table \ref{tab:Abl_DanceTrack} on DanceTrack dataset. This table presents a comparison of \text{TP-GMOT} with the existing fully-supervised MOT methods. To ensure a fair comparison, we have implemented a fully-supervised \text{MAC-SORT} method with YOLOX object detection. While our \text{MAC-SORT} with YOLOX achieves the best performance on HOTA, IDF1, it is noteworthy that \text{TP-GMOT}, which does not require any training data, demonstrates compatibility with SOTA fully-supervised MOT methods, even surpassing SORT, DeepSORT, and MOTDT. In this experiment, both general prompt and specific prompt are set as ``dancer''.

\begin{table}[!h]
\centering

\caption{Ablation study on \textit{hyper-parameters} on \textit{$\kappa_1$ for ($\mathcal{L}^M$) and  \textit{$\kappa_2$} for ($\mathcal{S}^M$)}.}
\resizebox{\linewidth}{!}{%
\begin{tabular}{l|ccc||l|ccc}
\toprule
\multicolumn{4}{c||}{\textbf{$\kappa_1$ for $\mathcal{L}^M$ ($\kappa_2 = 3$)}} & \multicolumn{4}{c}{\textbf{$\kappa_2$ for $\mathcal{S}^M$ ($\kappa_1 = 9$)}} \\ \hline
$\kappa_1$ & HOTA$\uparrow$   & MOTA$\uparrow$  & IDF1$\uparrow$    & $\kappa_2$ & HOTA$\uparrow$      &  MOTA$\uparrow$ & IDF1$\uparrow$   \\ \hline
1 & 57.86 & 67.02 & 69.85 & 1 & 58.07 & 66.97 & 70.26 \\
3 & 57.97 & 66.97 & 70.08 & 2 & 58.34 & 67.22 & 71.53 \\
6 & 58.08 & 66.98 & 70.19& 3 & 58.58 & 67.77 & 71.70\\
9 & 58.58 & 67.77 & 71.70& 4 & 58.45 & 67.27 & 71.60 \\
11 & 58.07 & 67.00 & 70.18& 5 & 58.44 & 67.27 & 71.58 \\

\bottomrule
\end{tabular}}
\label{tab:ablation_LSM}

\end{table}

\begin{table}[!h]
  \centering
  \setlength{\tabcolsep}{7pt}
  \renewcommand{\arraystretch}{1.1}

  \caption{\underline{Ablation study of $\theta$} in computing $W_{aaw}$ (Equation \ref{eq:theta}), on \textit{Refer-GMOT40} dataset.}
  \label{tab:gmot40}
  \resizebox{.9\linewidth}{!}{%
\begin{tabular}{l|cc|cc}
\toprule
\multirow{2}{*}{$\theta$} &  \multicolumn{2}{c|}{\textbf{Standard MOT metrics}}  & \multicolumn{2}{c}{\textbf{ID metrics}} \\ \cline{2-5} 
 & HOTA$\uparrow$           & MOTA$\uparrow$        & IDF1$\uparrow$           & IDR$\uparrow$            \\ \midrule  
22.5$^\circ$ & 57.54 & 66.82 &  69.29 &  64.16 \\
45 (default)$^\circ$ & \textbf{59.26} & \textbf{68.03 }& \textbf{70.86} &  \textbf{68.39} \\   
67.5$^\circ$ & 58.06 & 67.37 & 70.46 & 65.24 \\
80$^\circ$ &58.21   &65.81  &70.13   &68.30   \\
\bottomrule
\end{tabular}}
\label{tab:Abl_hyper-param}
\end{table}

\begin{table}[!h]
  \centering
  \caption{\underline{Ablation study on the impact of each modules} in our proposed \textbf{{TP-GMOT} }framework on \textit{Refer-GMOT40} dataset. SORT is a substitute when {MAC-SORT} is not used.}
\resizebox{\columnwidth}{!}{
\begin{tabular}{l|c|c|c||ccccc}
\toprule
\multirow{3}{*}{\textbf{Exp.}} & \multicolumn{2}{c|}{\textbf{{TP-OD}}} &  \multirow{2}{*}{\textbf{{MAC-SORT}}} & \multicolumn{5}{c}{\textbf{Standard MOT metrics}}\\ \cline{2-3} \cline{5-9}
& { {IE Strategy}}   & {{LSM Mechanism}} &  & HOTA$\uparrow$   & MOTA$\uparrow$  & IDs$\downarrow$           & DetA$\uparrow$           & AssA$\uparrow$    \\ \hline
\#0 & \xmark   &   \xmark  & \xmark (baseline) & 30.05          & 20.83             &  5,387 & 23.02          & 40.18    \\ \midrule                    
\#1 & \checkmark & \xmark & \xmark & 56.89  & 66.52 & 9,840 & 57.90 & 57.28 \\ 
\#2 &\xmark & \checkmark & \xmark &  56.15 & 66.05 & 9,088 & 56.98 & 56.70 \\ 
\#3 &\checkmark & \checkmark & \xmark  & 56.43 &  66.72 & 9,111 & 57.47 & 56.76 \\ 
\#4 &\checkmark & \checkmark &  \checkmark  &\textbf{58.37} &\textbf{67.68} &\textbf{5,898} &\textbf{58.32} &\textbf{59.82} \\
\bottomrule
\end{tabular}}
  \label{tab:ab_model}%
\end{table}%

\noindent
\underline{Investigation for choosing band size of LSM} We investigate the impact of varying the memory band size ($\kappa_1$ in  $\mathcal{L}^M$ and $\kappa_2$ in $\mathcal{S}^M$) in LSM strategy as presented in Table \ref{tab:ablation_LSM}.



\noindent
\underline{Effectiveness of proposed \text{MAC-SORT} on MOT task.} We assess \text{MAC-SORT}'s performance by conducting a comparison on the MOT20 dataset, as outlined in Table \ref{tb:Abl_MOT20}, focusing on the MOT task. To ensure a fair comparison, we disable certain ad-hoc settings that employ varying thresholds for individual sequences, an offline interpolation procedure, and internal weights for object detector. In this experiment, YOLOX object detector is used for all trackers to demonstrate the effectiveness of \text{MAC-SORT}.

\noindent
\underline{Hyper-param $\theta$.}
Table \ref{tab:Abl_hyper-param} shows ablation study of vector distance threshold $\theta$ as defined in Equation \ref{eq:theta}. The minor variations in performance demonstrate the robustness of our proposed $\mathtt{MA-SORT}$ when $\theta$ is varied within the range of [22.5$^\circ$, 80$^\circ$]. We select $\theta = 45$ in the reported results.

\noindent
\underline{Impact of each module in TP-GMOT.} Table \ref{tab:ab_model} illustrates the impact of TP-OD and MAC-SORT modules within the TP-GMOT framework. It is important to clarify that TP-OD is a framework implemented by our iGDINO, encompassing IE strategy and LSM mechanism. In the baseline, SORT is utilized as a substitute for the proposed MAC-SORT and VP-OD is used as objects detector.

\section{Conclusion \& Discussion}

\textbf{Conclusion.} In this study, we present \text{TP-GMOT}, a novel tracking framework capable of tracking diverse objects without relying on labeled tracking data. \text{TP-GMOT} offers two key contributions: (i) \text{TP-OD} framework with iGDINO for effective object detection using natural language descriptions and (ii) \text{MAC-SORT} for efficient tracking of visually similar objects within a broader context of generic objects. Beyond proposing \text{TP-GMOT}, we also introduce a new \text{Refer-GMOT} dataset. We have thoroughly assessed and demonstrated the efficacy and generalizability of \text{TP-GMOT}, not only in the GMOT task but also in the MOT task.

\textbf{Discussion.} The primary constraint in our framework lies in iGDINO's performance, which depend on pre-trained VLMs (GroundingDINO). For example, the GroundingDINO model, which incorporates horizontal flipping augmentation during training, presents a challenge for our \text{iGDINO} model in distinguishing between the concepts of ``left'' and ``right''. Nonetheless, we believe that this limitation can be effectively overcome by improvements, such as integrating more advanced data augmentation techniques during the pre-training phase. In addition to GroundingDINO, it is important to recognize the rich diversity of VLMs available in the field, which opens up exciting avenues for deeper exploration. In our current work, we have implemented \text{TP-GMOT} exclusively using only textual description \texttt{caption}. Nevertheless, our \text{Refer-GMOT} dataset offers additional information which hold great potential for further research, particularly in the context prompt engineering. Exploring these additional aspects of \text{Refer-GMOT} dataset could lead to enhanced object tracking capabilities as well as other fields such as surveillance, robotics, and animal welfare. We expect our work to inspire future research in the unexplored realm of open-vocabulary, open-world, MOT/GMOT paradigms.



\bibliography{m722}

\begin{thebibliography}{60}
\providecommand{\natexlab}[1]{#1}
\providecommand{\url}[1]{\texttt{#1}}
\expandafter\ifx\csname urlstyle\endcsname\relax
  \providecommand{\doi}[1]{doi: #1}\else
  \providecommand{\doi}{doi: \begingroup \urlstyle{rm}\Url}\fi

\bibitem[Aharon et~al.(2022)Aharon, Orfaig, and Bobrovsky]{aharon2022bot}
N.~Aharon, R.~Orfaig, and B.-Z. Bobrovsky.
\newblock Bot-sort: Robust associations multi-pedestrian tracking.
\newblock \emph{arXiv preprint arXiv:2206.14651}, 2022.

\bibitem[Bai et~al.(2021)Bai, Cheng, Chu, Liu, Zhang, and Ling]{bai2021gmot}
H.~Bai, W.~Cheng, P.~Chu, J.~Liu, K.~Zhang, and H.~Ling.
\newblock Gmot-40: A benchmark for generic multiple object tracking.
\newblock In \emph{CVPR}, pages 6719--6728, 2021.

\bibitem[Bernardin and Stiefelhagen(2008)]{Bernardin2008mota}
K.~Bernardin and R.~Stiefelhagen.
\newblock Evaluating multiple object tracking performance: The clear mot metrics.
\newblock \emph{EURASIP Journal on Image and Video Processing}, 2008.

\bibitem[Bewley et~al.(2016)Bewley, Ge, Ott, Ramos, and Upcroft]{bewley2016simple}
A.~Bewley, Z.~Ge, L.~Ott, F.~Ramos, and B.~Upcroft.
\newblock Simple online and realtime tracking.
\newblock In \emph{ICIP}, pages 3464--3468. IEEE, 2016.

\bibitem[Bras{\'o} and Leal-Taix{\'e}(2020)]{braso2020learning}
G.~Bras{\'o} and L.~Leal-Taix{\'e}.
\newblock Learning a neural solver for multiple object tracking.
\newblock In \emph{CVPR}, pages 6247--6257, 2020.

\bibitem[Cai et~al.(2022{\natexlab{a}})Cai, Xu, Li, Xiong, Xia, Tu, and Soatto]{cai2022memot}
J.~Cai, M.~Xu, W.~Li, Y.~Xiong, W.~Xia, Z.~Tu, and S.~Soatto.
\newblock Memot: Multi-object tracking with memory.
\newblock In \emph{CVPR}, pages 8090--8100, 2022{\natexlab{a}}.

\bibitem[Cai et~al.(2022{\natexlab{b}})Cai, Kwon, Ravichandran, Bas, Tu, Bhotika, and Soatto]{cai2022x}
Z.~Cai, G.~Kwon, A.~Ravichandran, E.~Bas, Z.~Tu, R.~Bhotika, and S.~Soatto.
\newblock X-detr: A versatile architecture for instance-wise vision-language tasks.
\newblock \emph{ECCV}, 2022{\natexlab{b}}.

\bibitem[Cao et~al.(2023)Cao, Pang, Weng, Khirodkar, and Kitani]{cao2023observation}
J.~Cao, J.~Pang, X.~Weng, R.~Khirodkar, and K.~Kitani.
\newblock Observation-centric sort: Rethinking sort for robust multi-object tracking.
\newblock In \emph{CVPR}, pages 9686--9696, 2023.

\bibitem[Chan et~al.(2022)Chan, Jia, Zhou, Bai, Chen, and Zhang]{CHAN2022108793}
S.~Chan, Y.~Jia, X.~Zhou, C.~Bai, S.~Chen, and X.~Zhang.
\newblock Online multiple object tracking using joint detection and embedding network.
\newblock \emph{Pattern Recognition}, 130:\penalty0 108793, 2022.

\bibitem[Chen et~al.(2018)Chen, Ai, Zhuang, and Shang]{Chen2018RealTimeMP}
L.~Chen, H.~Ai, Z.~Zhuang, and C.~Shang.
\newblock Real-time multiple people tracking with deeply learned candidate selection and person re-identification.
\newblock \emph{ICME}, 2018.

\bibitem[Cui et~al.(2023)Cui, Zeng, Zhao, Yang, Wu, and Wang]{cui2023sportsmot}
Y.~Cui, C.~Zeng, X.~Zhao, Y.~Yang, G.~Wu, and L.~Wang.
\newblock Sportsmot: A large multi-object tracking dataset in multiple sports scenes.
\newblock \emph{arXiv preprint arXiv:2304.05170}, 2023.

\bibitem[Dave et~al.(2020)Dave, Khurana, Tokmakov, Schmid, and Ramanan]{dave2020tao}
A.~Dave, T.~Khurana, P.~Tokmakov, C.~Schmid, and D.~Ramanan.
\newblock Tao: A large-scale benchmark for tracking any object.
\newblock In \emph{ECCV}, pages 436--454. Springer, 2020.

\bibitem[Dendorfer et~al.(2020)Dendorfer, Rezatofighi, Milan, Shi, Cremers, Reid, Roth, Schindler, and Leal-Taixe]{dendorfer2020mot20}
P.~Dendorfer, H.~Rezatofighi, A.~Milan, J.~Shi, D.~Cremers, I.~Reid, S.~Roth, K.~Schindler, and L.~Leal-Taixe.
\newblock Mot20: A benchmark for multi object tracking in crowded scenes.
\newblock \emph{arXiv preprint arXiv:2003.09003}, 2020.

\bibitem[Ding et~al.(2022)Ding, Wang, and Tu]{ding2022open}
Z.~Ding, J.~Wang, and Z.~Tu.
\newblock Open-vocabulary panoptic segmentation with maskclip.
\newblock \emph{arXiv preprint arXiv:2208.08984}, 2022.

\bibitem[Du et~al.(2023)Du, Zhao, Song, Zhao, Su, Gong, and Meng]{du2023strongsort}
Y.~Du, Z.~Zhao, Y.~Song, Y.~Zhao, F.~Su, T.~Gong, and H.~Meng.
\newblock Strongsort: Make deepsort great again.
\newblock \emph{IEEE TMM}, 2023.

\bibitem[Fischer et~al.(2023)Fischer, Huang, Pang, Qiu, Chen, Darrell, and Yu]{fischer2023qdtrack}
T.~Fischer, T.~E. Huang, J.~Pang, L.~Qiu, H.~Chen, T.~Darrell, and F.~Yu.
\newblock Qdtrack: Quasi-dense similarity learning for appearance-only multiple object tracking.
\newblock \emph{IEEE TPAMI}, 2023.

\bibitem[Ge et~al.(2021)Ge, Liu, Wang, Li, and Sun]{yolox2021}
Z.~Ge, S.~Liu, F.~Wang, Z.~Li, and J.~Sun.
\newblock Yolox: Exceeding yolo series in 2021.
\newblock \emph{arXiv preprint arXiv:2107.08430}, 2021.

\bibitem[Ghiasi et~al.(2022)Ghiasi, Gu, Cui, and Lin]{ghiasi2022open}
G.~Ghiasi, X.~Gu, Y.~Cui, and T.-Y. Lin.
\newblock Open-vocabulary image segmentation.
\newblock \emph{ECCV}, 2022.

\bibitem[Gu et~al.(2022)Gu, Lin, Kuo, and Cui]{gu2022open}
X.~Gu, T.-Y. Lin, W.~Kuo, and Y.~Cui.
\newblock Open-vocabulary object detection via vision and language knowledge distillation.
\newblock \emph{ICLR}, 2022.

\bibitem[Huang et~al.(2020)Huang, Zhao, and Huang]{huang2020globaltrack}
L.~Huang, X.~Zhao, and K.~Huang.
\newblock Globaltrack: A simple and strong baseline for long-term tracking.
\newblock In \emph{AAAI}, 2020.

\bibitem[Jia et~al.(2021)Jia, Yang, Xia, Chen, Parekh, Pham, Le, Sung, Li, and Duerig]{jia2021scaling}
C.~Jia, Y.~Yang, Y.~Xia, Y.-T. Chen, Z.~Parekh, H.~Pham, Q.~V. Le, Y.~Sung, Z.~Li, and T.~Duerig.
\newblock Scaling up visual and vision-language representation learning with noisy text supervision.
\newblock In \emph{ICLR}, pages 4904--4916. PMLR, 2021.

\bibitem[Leal-Taix{\'e} et~al.(2016)Leal-Taix{\'e}, Canton-Ferrer, and Schindler]{leal2016learning}
L.~Leal-Taix{\'e}, C.~Canton-Ferrer, and K.~Schindler.
\newblock Learning by tracking: Siamese cnn for robust target association.
\newblock In \emph{CVPRW}, pages 33--40, 2016.

\bibitem[Li et~al.(2022{\natexlab{a}})Li, Weinberger, Belongie, Koltun, and Ranftl]{li2022language}
B.~Li, K.~Q. Weinberger, S.~Belongie, V.~Koltun, and R.~Ranftl.
\newblock Language-driven semantic segmentation.
\newblock \emph{ICLR}, 2022{\natexlab{a}}.

\bibitem[Li et~al.(2022{\natexlab{b}})Li, Zhang, Zhang, Yang, Li, Zhong, Wang, Yuan, Zhang, Hwang, Chang, and Gao]{li2022grounded}
L.~H. Li, P.~Zhang, H.~Zhang, J.~Yang, C.~Li, Y.~Zhong, L.~Wang, L.~Yuan, L.~Zhang, J.-N. Hwang, K.-W. Chang, and J.~Gao.
\newblock Grounded language-image pre-training.
\newblock In \emph{CVPR}, pages 10965--10975, 2022{\natexlab{b}}.

\bibitem[Li et~al.(2023{\natexlab{a}})Li, Fischer, Ke, Ding, Danelljan, and Yu]{li2023ovtrack}
S.~Li, T.~Fischer, L.~Ke, H.~Ding, M.~Danelljan, and F.~Yu.
\newblock Ovtrack: Open-vocabulary multiple object tracking.
\newblock In \emph{Proceedings of the IEEE/CVF CVPR}, pages 5567--5577, 2023{\natexlab{a}}.

\bibitem[Li et~al.(2023{\natexlab{b}})Li, Liu, Wu, Mu, Yang, Gao, Li, and Lee]{li2023gligen}
Y.~Li, H.~Liu, Q.~Wu, F.~Mu, J.~Yang, J.~Gao, C.~Li, and Y.~J. Lee.
\newblock Gligen: Open-set grounded text-to-image generation.
\newblock In \emph{CVPR}, pages 22511--22521, 2023{\natexlab{b}}.

\bibitem[Liang et~al.(2022)Liang, Zhang, Lu, Zhou, Li, Ye, and Zou]{liang2022cstrack}
C.~Liang, Z.~Zhang, Y.~Lu, X.~Zhou, B.~Li, X.~Ye, and J.~Zou.
\newblock Rethinking the competition between detection and reid in multi-object tracking.
\newblock \emph{IEEE TIP}, 2022.

\bibitem[Liu et~al.(2019)Liu, Liu, Ren, He, and Sun]{liu2019aligning}
F.~Liu, Y.~Liu, X.~Ren, X.~He, and X.~Sun.
\newblock Aligning visual regions and textual concepts for semantic-grounded image representations.
\newblock \emph{Advances in Neural Information Processing Systems}, 32, 2019.

\bibitem[Liu et~al.(2023{\natexlab{a}})Liu, Li, Jiang, and Fu]{liu2023siamesedetr}
Q.~Liu, Y.~Li, Y.~Jiang, and Y.~Fu.
\newblock Siamese-detr for generic multi-object tracking, 2023{\natexlab{a}}.

\bibitem[Liu et~al.(2023{\natexlab{b}})Liu, Zeng, Ren, Li, Zhang, Yang, Jiang, Li, Yang, Su, Zhu, and Zhang]{liu2023grounding}
S.~Liu, Z.~Zeng, T.~Ren, F.~Li, H.~Zhang, J.~Yang, Q.~Jiang, C.~Li, J.~Yang, H.~Su, J.~Zhu, and L.~Zhang.
\newblock Grounding dino: Marrying dino with grounded pre-training for open-set object detection.
\newblock \emph{arXiv preprint arXiv:2303.05499}, 2023{\natexlab{b}}.

\bibitem[Luiten et~al.(2020)Luiten, Osep, Dendorfer, Torr, Geiger, Leal-Taixé, and Leibe]{Luiten_2020hota}
J.~Luiten, A.~Osep, P.~Dendorfer, P.~Torr, A.~Geiger, L.~Leal-Taixé, and B.~Leibe.
\newblock Hota: A higher order metric for evaluating multi-object tracking.
\newblock \emph{International Journal of Computer Vision}, 2020.

\bibitem[Luo and Kim(2013)]{luo2013generic}
W.~Luo and T.-K. Kim.
\newblock Generic object crowd tracking by multi-task learning.
\newblock In \emph{BMVC}, volume~1, page~3, 2013.

\bibitem[Luo et~al.(2014)Luo, Kim, Stenger, Zhao, and Cipolla]{luo2014bi}
W.~Luo, T.-K. Kim, B.~Stenger, X.~Zhao, and R.~Cipolla.
\newblock Bi-label propagation for generic multiple object tracking.
\newblock In \emph{CVPR}, pages 1290--1297, 2014.

\bibitem[Maggiolino et~al.(2023)Maggiolino, Ahmad, Cao, and Kitani]{maggiolino2023deep}
G.~Maggiolino, A.~Ahmad, J.~Cao, and K.~Kitani.
\newblock Deep oc-sort: Multi-pedestrian tracking by adaptive re-identification.
\newblock \emph{arXiv preprint arXiv:2302.11813}, 2023.

\bibitem[Meinhardt et~al.(2022)Meinhardt, Kirillov, Leal-Taixe, and Feichtenhofer]{meinhardt2022trackformer}
T.~Meinhardt, A.~Kirillov, L.~Leal-Taixe, and C.~Feichtenhofer.
\newblock Trackformer: Multi-object tracking with transformers.
\newblock In \emph{CVPR}, pages 8844--8854, 2022.

\bibitem[Milan et~al.(2016)Milan, Leal-Taixe, Reid, Roth, and Schindler]{milan2016mot16}
A.~Milan, L.~Leal-Taixe, I.~Reid, S.~Roth, and K.~Schindler.
\newblock Mot16: A benchmark for multi-object tracking.
\newblock \emph{arXiv preprint arXiv:1603.00831}, 2016.

\bibitem[Minderer et~al.(2022)Minderer, Gritsenko, Stone, Neumann, Weissenborn, Alexey~Dosovitskiy, Arnab, Dehghani, Shen, Wang, Zhai, Kipf, and Houlsby]{minderer2022simple}
M.~Minderer, A.~Gritsenko, A.~Stone, M.~Neumann, D.~Weissenborn, A.~M. Alexey~Dosovitskiy, A.~Arnab, M.~Dehghani, Z.~Shen, X.~Wang, X.~Zhai, T.~Kipf, and N.~Houlsby.
\newblock Simple open-vocabulary object detection with vision transformers.
\newblock \emph{ECCV}, 2022.

\bibitem[Pang et~al.(2021)Pang, Qiu, Li, Chen, Li, Darrell, and Yu]{pang2021quasi}
J.~Pang, L.~Qiu, X.~Li, H.~Chen, Q.~Li, T.~Darrell, and F.~Yu.
\newblock Quasi-dense similarity learning for multiple object tracking.
\newblock In \emph{CVPR}, pages 164--173, 2021.

\bibitem[Radford et~al.(2021)Radford, Kim, Hallacy, Ramesh, Goh, Agarwal, Sastry, Askell, Mishkin, Clark, Krueger, and Sutskever]{radford2021learning}
A.~Radford, J.~W. Kim, C.~Hallacy, A.~Ramesh, G.~Goh, S.~Agarwal, G.~Sastry, A.~Askell, P.~Mishkin, J.~Clark, G.~Krueger, and I.~Sutskever.
\newblock Learning transferable visual models from natural language supervision.
\newblock In \emph{ICML}, pages 8748--8763. PMLR, 2021.

\bibitem[Rao et~al.(2022)Rao, Zhao, Chen, Tang, Zhu, Huang, Zhou, and Lu]{rao2022denseclip}
Y.~Rao, W.~Zhao, G.~Chen, Y.~Tang, Z.~Zhu, G.~Huang, J.~Zhou, and J.~Lu.
\newblock Denseclip: Language-guided dense prediction with context-aware prompting.
\newblock In \emph{CVPR}, pages 18082--18091, 2022.

\bibitem[Ren et~al.(2015)Ren, He, Girshick, and Sun]{ren2015faster}
S.~Ren, K.~He, R.~Girshick, and J.~Sun.
\newblock Faster r-cnn: Towards real-time object detection with region proposal networks.
\newblock In \emph{NIPS}, pages 91--99, 2015.

\bibitem[Ristani et~al.(2016)Ristani, Solera, Zou, Cucchiara, and Tomasi]{Ristani2016idf1}
E.~Ristani, F.~Solera, R.~Zou, R.~Cucchiara, and C.~Tomasi.
\newblock \emph{Performance Measures and a Data Set for Multi-target, Multi-camera Tracking}.
\newblock Springer International Publishing, 2016.

\bibitem[Sun et~al.(2020{\natexlab{a}})Sun, Cao, Jiang, Zhang, Xie, Yuan, Wang, and Luo]{transtrack}
P.~Sun, J.~Cao, Y.~Jiang, R.~Zhang, E.~Xie, Z.~Yuan, C.~Wang, and P.~Luo.
\newblock Transtrack: Multiple-object tracking with transformer.
\newblock \emph{arXiv preprint arXiv: 2012.15460}, 2020{\natexlab{a}}.

\bibitem[Sun et~al.(2022)Sun, Cao, Jiang, Yuan, Bai, Kitani, and Luo]{sun2022dancetrack}
P.~Sun, J.~Cao, Y.~Jiang, Z.~Yuan, S.~Bai, K.~Kitani, and P.~Luo.
\newblock Dancetrack: Multi-object tracking in uniform appearance and diverse motion.
\newblock In \emph{CVPR}, pages 20993--21002, 2022.

\bibitem[Sun et~al.(2020{\natexlab{b}})Sun, Akhtar, Song, Song, Mian, and Shah]{sun2020simultaneous}
S.~Sun, N.~Akhtar, X.~Song, H.~Song, A.~Mian, and M.~Shah.
\newblock Simultaneous detection and tracking with motion modelling for multiple object tracking.
\newblock In \emph{ECCV}, pages 626--643. Springer, 2020{\natexlab{b}}.

\bibitem[Wang et~al.(2020)Wang, Kitani, and Weng]{Wang2020_GSDT}
Y.~Wang, K.~Kitani, and X.~Weng.
\newblock {Joint Object Detection and Multi-Object Tracking with Graph Neural Networks}.
\newblock \emph{arXiv:2006.13164}, 2020.

\bibitem[Wojke et~al.(2017)Wojke, Bewley, and Paulus]{wojke2017simple}
N.~Wojke, A.~Bewley, and D.~Paulus.
\newblock Simple online and realtime tracking with a deep association metric.
\newblock In \emph{ICIP}, pages 3645--3649. IEEE, 2017.

\bibitem[Wu et~al.(2023)Wu, Han, Wang, Dong, Zhang, and Shen]{wu2023referring}
D.~Wu, W.~Han, T.~Wang, X.~Dong, X.~Zhang, and J.~Shen.
\newblock Referring multi-object tracking.
\newblock In \emph{CVPR}, pages 14633--14642, 2023.

\bibitem[Wu et~al.(2021)Wu, Cao, Song, Wang, Yang, and Yuan]{wu2021track}
J.~Wu, J.~Cao, L.~Song, Y.~Wang, M.~Yang, and J.~Yuan.
\newblock Track to detect and segment: An online multi-object tracker.
\newblock In \emph{CVPR}, pages 12352--12361, 2021.

\bibitem[Yan et~al.(2022)Yan, Jiang, Sun, Wang, Yuan, Luo, and Lu]{yan2022towards}
B.~Yan, Y.~Jiang, P.~Sun, D.~Wang, Z.~Yuan, P.~Luo, and H.~Lu.
\newblock Towards grand unification of object tracking.
\newblock In \emph{ECCV}, 2022.

\bibitem[Yang et~al.(2022)Yang, Li, Zhang, Xiao, Liu, Yuan, and Gao]{yang2022unified}
J.~Yang, C.~Li, P.~Zhang, B.~Xiao, C.~Liu, L.~Yuan, and J.~Gao.
\newblock Unified contrastive learning in image-text-label space.
\newblock In \emph{CVPR}, pages 19163--19173, 2022.

\bibitem[Zeng et~al.(2022)Zeng, Dong, Zhang, Wang, Zhang, and Wei]{zeng2022motr}
F.~Zeng, B.~Dong, Y.~Zhang, T.~Wang, X.~Zhang, and Y.~Wei.
\newblock Motr: End-to-end multiple-object tracking with transformer.
\newblock In \emph{ECCV}, pages 659--675. Springer, 2022.

\bibitem[Zhang et~al.(2022{\natexlab{a}})Zhang, Zhang, Hu, Chen, Li, Dai, Wang, Yuan, Hwang, and Gao]{zhang2022glipv2}
H.~Zhang, P.~Zhang, X.~Hu, Y.-C. Chen, L.~H. Li, X.~Dai, L.~Wang, L.~Yuan, J.-N. Hwang, and J.~Gao.
\newblock Glipv2: Unifying localization and vision-language understanding.
\newblock \emph{NIPS}, 2022{\natexlab{a}}.

\bibitem[Zhang et~al.(2022{\natexlab{b}})Zhang, Gao, Xiao, and Fan]{zhang2022animaltrack}
L.~Zhang, J.~Gao, Z.~Xiao, and H.~Fan.
\newblock Animaltrack: A benchmark for multi-animal tracking in the wild.
\newblock \emph{IJCV}, pages 1--18, 2022{\natexlab{b}}.

\bibitem[Zhang et~al.(2021{\natexlab{a}})Zhang, Shi, Tang, Xiao, Yu, and Zhuang]{zhang2021consensus}
W.~Zhang, H.~Shi, S.~Tang, J.~Xiao, Q.~Yu, and Y.~Zhuang.
\newblock Consensus graph representation learning for better grounded image captioning.
\newblock In \emph{AAAI}, volume~35, pages 3394--3402, 2021{\natexlab{a}}.

\bibitem[Zhang et~al.(2021{\natexlab{b}})Zhang, Wang, Wang, Zeng, and Liu]{zhang2021fairmot}
Y.~Zhang, C.~Wang, X.~Wang, W.~Zeng, and W.~Liu.
\newblock Fairmot: On the fairness of detection and re-identification in multiple object tracking.
\newblock \emph{IJCV}, 129:\penalty0 3069--3087, 2021{\natexlab{b}}.

\bibitem[Zhang et~al.(2022{\natexlab{c}})Zhang, Sun, Jiang, Yu, Weng, Yuan, Luo, Liu, and Wang]{zhang2021bytetrack}
Y.~Zhang, P.~Sun, Y.~Jiang, D.~Yu, F.~Weng, Z.~Yuan, P.~Luo, W.~Liu, and X.~Wang.
\newblock Bytetrack: Multi-object tracking by associating every detection box.
\newblock In \emph{ECCV}, 2022{\natexlab{c}}.

\bibitem[Zhang et~al.(2023)Zhang, Wang, and Zhang]{zhang2023motrv2}
Y.~Zhang, T.~Wang, and X.~Zhang.
\newblock Motrv2: Bootstrapping end-to-end multi-object tracking by pretrained object detectors.
\newblock In \emph{CVPR}, pages 22056--22065, 2023.

\bibitem[Zhong et~al.(2022)Zhong, Yang, Zhang, Li, Codella, Li, Zhou, Dai, Yuan, Li, and Gao]{zhong2022regionclip}
Y.~Zhong, J.~Yang, P.~Zhang, C.~Li, N.~Codella, L.~H. Li, L.~Zhou, X.~Dai, L.~Yuan, Y.~Li, and J.~Gao.
\newblock Regionclip: Region-based language-image pretraining.
\newblock In \emph{CVPR}, pages 16793--16803, 2022.

\bibitem[Zhou et~al.(2020)Zhou, Koltun, and Kr{\"a}henb{\"u}hl]{zhou2020tracking}
X.~Zhou, V.~Koltun, and P.~Kr{\"a}henb{\"u}hl.
\newblock Tracking objects as points.
\newblock In \emph{ECCV}, pages 474--490, 2020.

\end{thebibliography}

\end{document}


\begin{frontmatter}
\paperid{722} 


\title{TP-GMOT: Tracking Generic Multiple Object by Textual Prompt with Motion-Appearance Cost (MAC) SORT
Supplementary}


\author{\fnms{Duy Le Dinh Anh$^{1, 2}$, Kim Hoang Tran$^{1,2}$, {\bf Ngan Hoang Le$^2$}}\\

    $^1$ FPT Software AI Center, Vietnam \\
    $^2$ Department of Computer Science, University of Arkansas, USA}

\end{frontmatter}

Within this supplementary document, our goal is to provide a comprehensive overview of study. We initiate with a thorough analysis of our \texttt{Refer-GMOT} dataset in Section \ref{sec:dataset}, where we provide details of data labeling process in Section \ref{sec:anno_proce}, and present insights from our analysis in Section \ref{sec:stat_in}. We then further offer ablation studies and qualitative results in Section \ref{sec:ablation} to showcase the effectiveness of our \texttt{TP-GMOT} framework consisting of the proposed \texttt{TP-OD} for object detection and \texttt{MAC-SORT} for object association.


\section{\texttt{Refer-GMOT} dataset}
\label{sec:dataset}

\subsection{Data Annotation Procedure}
\label{sec:anno_proce}
\begin{figure*}[!t]
    \centering    \includegraphics[width=0.9\textwidth]{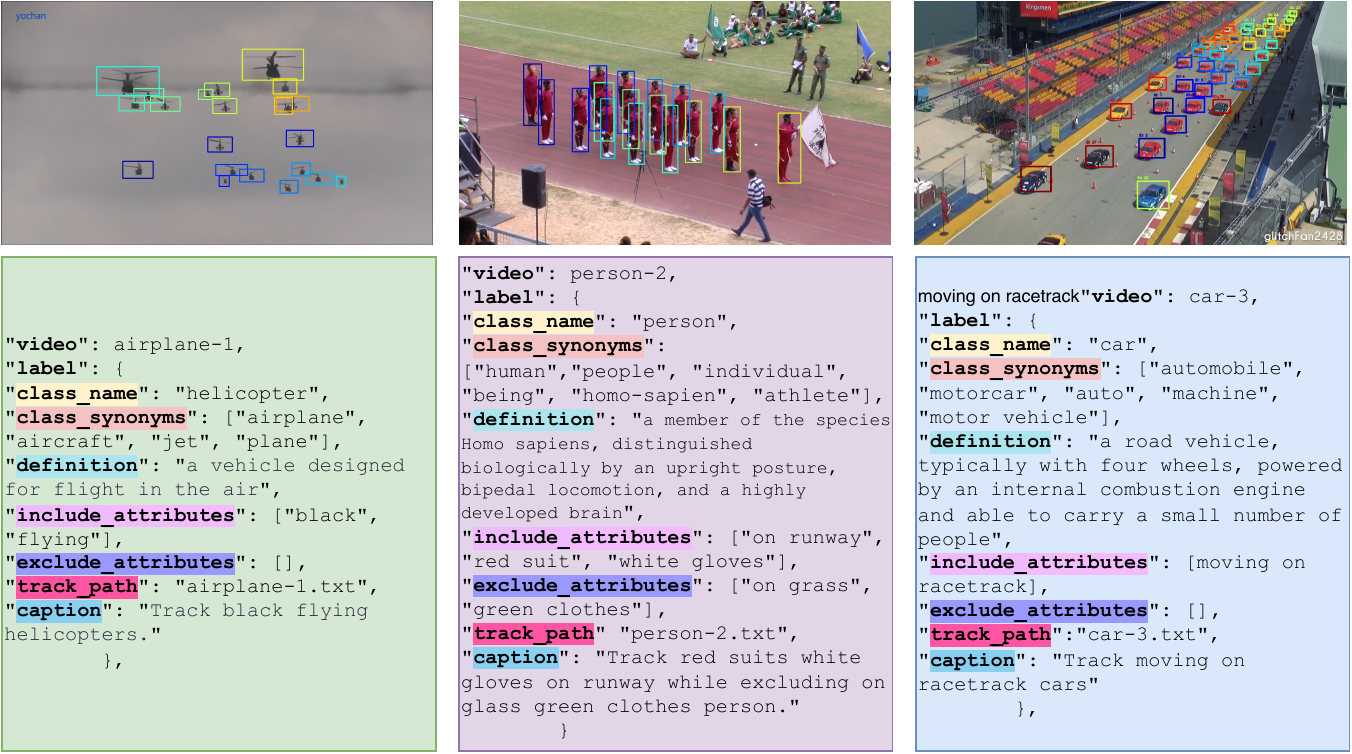}
    \caption{Demonstrations of textual description in our dataset.}
    \label{fig:dataset}
\end{figure*}

Instead of acquiring new videos, our approach focuses on enhancing existing datasets, such as GMOT40 and AnimalTrack by adding textual descriptions. As a result, we conduct a \texttt{Refer-GMOT} dataset, which includes \texttt{Refer-GMOT40} and \texttt{Refer-Animal} datasets. These datasets, with their varied categories, are ideal for testing the generalization and robustness of tracking methods. The \texttt{Refer-GMOT} dataset has been thoroughly annotated by a team of four expert annotators who manually marked the data. This was followed by a rigorous double-checking process to ensure the annotations were both consistent and precise. Some examples of data annotation in the proposed \texttt{Ref-GMOT} dataset is given in Figure \ref{fig:dataset}.

\subsection{Statistical Insights of \texttt{Refer-GMOT}}
\label{sec:stat_in}

\begin{table}[!thb]
    \setlength{\tabcolsep}{8pt}
    \renewcommand{\arraystretch}{1.0}
    \centering
    \caption{Statistics insights of \texttt{Refer-GMOT}. \# denotes the quantity of the respective items.}
    \begin{tabular}{lrrr}
        \toprule
        \textbf{Class Name} & \# \textbf{Frames} & \# \textbf{Objects} & \# \textbf{Boxes} \\
        \midrule
        ant & 302 & 72 & 4668 \\
        airplane & 1181 & 81 & 23307 \\
        balloon & 3218 & 557 & 83430 \\
        ball & 729 & 181 & 20399 \\
        bee & 357 & 225 & 9439 \\
        bird & 3462 & 307 & 70972 \\
        boat & 1472 & 143 & 29491 \\
        car & 2023 & 221 & 33893 \\
        chicken & 7615 & 226 & 54710 \\
        deer & 2762 & 222 & 53339 \\
        dolphin & 1718 & 167 & 31112 \\
        duck & 4851 & 277 & 79416 \\
        fish & 569 & 291 & 21922 \\
        goose & 1773 & 195 & 33120 \\
        horse & 6556 & 281 & 69524 \\
        penguin & 1844 & 137 & 30312 \\
        person & 1040 & 106 & 22650 \\
        pig & 1531 & 151 & 32273 \\
        rabbit & 1558 & 313 & 33961 \\
        stock & 1128 & 143 & 34058 \\
        zebra & 1378 & 99 & 22331 \\
        \bottomrule
    \end{tabular}
    \label{tab:data_for_each_class}
\end{table}

Our \texttt{Refer-GMOT} dataset showcases substantial variation in the occurrence of critical elements, including the number of frames, objects, and bounding boxes, across different categories. Each category possesses unique characteristics, highlighting the diverse nature of our dataset, as illustrated in Table \ref{tab:data_for_each_class}. These categories encompass a wide range of object types and environments, spanning from humans to vehicles, animals to insects, and indoor settings to farms and open skies.

\section{Additional Ablation Experiments}
\label{sec:ablation}
\noindent

\subsection{Effectiveness of \texttt{TP-OD}}

\begin{figure*}[!t]
    \centering    \includegraphics[width=0.9\textwidth]{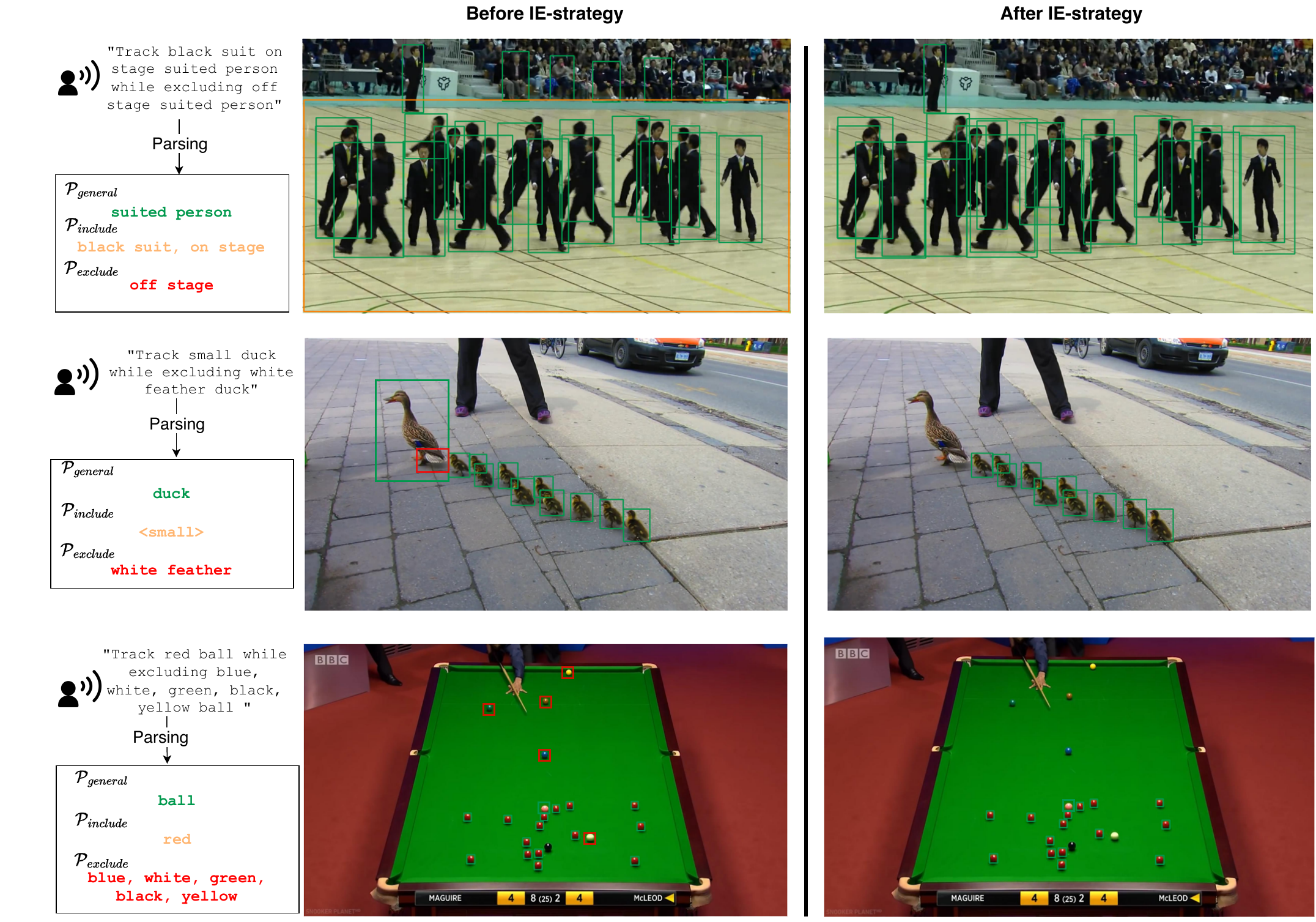}
    \caption{Examples to illustrate the efficacy of IE-Strategy. Left: Output from pre-trained VLM. Right: Output from IE-Strategy.}
    \label{fig:iewp}
\end{figure*}

Both the IE strategy and LSM mechanism within the \texttt{TP-OD} framework play crucial roles in accurately detecting objects, as illustrated in Figures \ref{fig:iewp}. Both serve as a filter to increase true positives (TPs) and reduce false positives (FPs). While the IE strategy focuses on verifying specific attributes of objects, the LSM mechanism acts as a secondary filter to recover TPs that the IE strategy may have incorrectly dismissed by cross-referencing them with the highest confidence detections from the memory bank. Together, these systems offer complementary validation methods, each reinforcing the other to ensure consistent and reliable performance.

\begin{figure*}[!t]
    \centering    \includegraphics[width=0.9\textwidth]{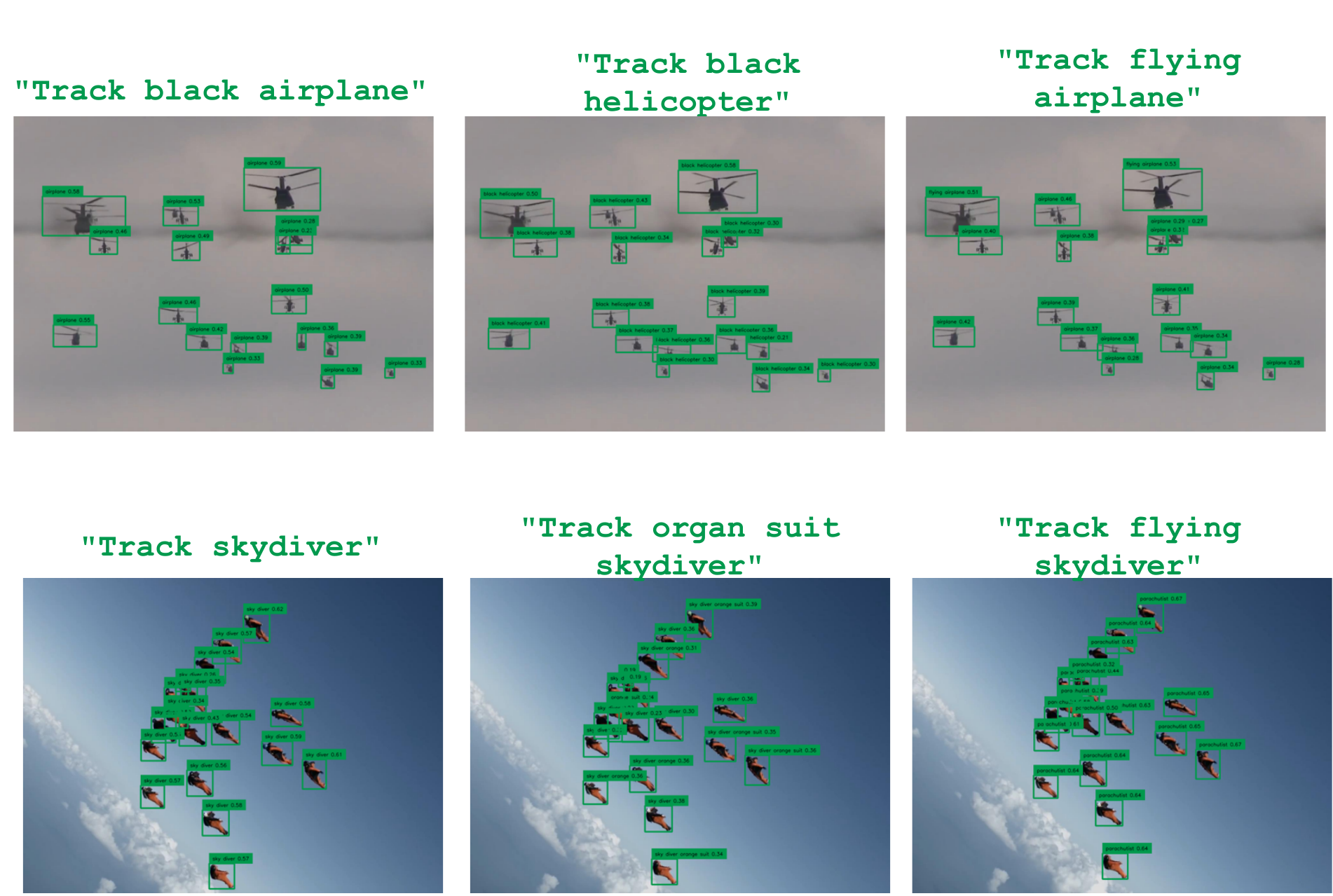}
    \caption{An illustration of flexibility and resilience of our proposed \texttt{TP-OD}. From left to right: object detection by our \texttt{TP-OD} with different synonym input prompts.}     
    \label{fig:detection}
\end{figure*}

We further demonstrate the flexibility of our object detection system \texttt{CSOD} in detecting objects with various synonym input prompts, as depicted in Figure \ref{fig:detection}.

\subsection{Ablation study on MAC-SORT}

To begin, let's revisit the cost matrix $C$ defined in \texttt{MAC-SORT} (as in Equation 11 in the main paper), outlined as below. It contains three components corresponding to IoU cost matrix, consistency between the directions, and visual apperance. Table \ref{tab:mac-sort} illustrates the influence of each component within \texttt{MAC-SORT} to tracking performance on \texttt{Refer-GMOT40} dataset.  In this table, we employ the same \texttt{TP-OD} for object detection.


\begin{table*}[!t]
  \centering

  \caption{Ablation study conducted on \texttt{Refer-GMOT40} dataset to showcase the impact of each component within the proposed \texttt{MAC-SORT}.}
\resizebox{0.95\textwidth}{!}{%
\begin{tabular}{ccc|ccccccc|ccc}
\toprule
{IoU}   & {Direction} &  \multirow{1}{*}{Appearance}& \multicolumn{7}{c|}{\textbf{Standard MOT metrics}}                                                                                      & \multicolumn{3}{c}{\textbf{ID metrics}}                   \\ \cline{4-13} 
Cost & Cost & Cost  & HOTA$\uparrow$   & MOTA$\uparrow$           & Pre$\uparrow$             & Re$\uparrow$             & TP$\uparrow$           & MT$\uparrow$                 & ML$\downarrow$           & IDF1$\uparrow$           & IDP$\uparrow$            & IDR$\uparrow$            \\ \hline
\checkmark & \xmark & \checkmark & 57.79 & 67.11 & 87.69  & 81.58 & 209,132 & 1,108 & 145 & 69.86 & 75.76 & 64.82  \\ \hline
\checkmark & \checkmark & \xmark & 57.71 & 67.07 & 87.69  & 81.58 & 209,132 & 1,107 &  145 & 69.79 & 75.67 & 64.75  \\ \hline
\checkmark & \checkmark &  \checkmark  & 58.58 & 67.77 & 87.69 & 81.58 & 209,132 & 1,237 &  140 & 71.70 & 73.51 & 68.39  \\
\bottomrule
\end{tabular}}
\label{tab:mac-sort}%
\end{table*}%